\documentclass[11pt]{article}
\usepackage[preprint]{acl}   

\usepackage{times}
\usepackage{latexsym}

\usepackage[T1]{fontenc}
\usepackage{tikz}

\usepackage[utf8]{inputenc}

\usepackage{microtype}

\usepackage{inconsolata}

\usepackage{graphicx}

\definecolor{darkgreen}{RGB}{0,120,0} 

\usepackage{dingbat}   
\usepackage{pifont}

\usepackage{geometry} 
\usepackage{multicol} 
\usepackage{float} 
\usepackage{booktabs}
\usepackage{amssymb}

\usepackage{array}
\usepackage{hyperref}
\usepackage{xcolor,pifont}

\usepackage{url}
\usepackage{graphicx} 
\usepackage{wrapfig} 
\usepackage{bbm}
\usepackage{booktabs}        
\usepackage{makecell}        
\usepackage{arydshln}        
\usepackage[table]{xcolor}   
\usepackage{subcaption}
\usepackage{amsmath}         
\usepackage{caption}
\usepackage{enumitem}
\usepackage{multirow}
\usepackage[utf8]{inputenc}
\usepackage{float} 
\usepackage{enumitem}
\usepackage{xcolor}
\usepackage[most]{tcolorbox}
\usepackage{listings}
\usepackage{fancyvrb}
\tcbuselibrary{breakable}
\usepackage{capt-of} 
\usepackage{wasysym}  

\newcommand{\cmark}{\CIRCLE}      
\newcommand{\pmark}{\LEFTcircle}  
\newcommand{\xmark}{\Circle}      

\title{Quantifying LLM Biases Across Instruction Boundary \\
in Mixed Question Forms}

\author{
  Zipeng Ling$^{2*}$,
  Shuliang Liu$^{1*}$,
  Yuehao Tang$^{1*}$,
  Chen Huang$^{3*}$,
  Gaoyang Jiang$^{3}$\\
  \textbf{
  Shenghong Fu$^{4}$,
  Junqi Yang$^{3}$,
  Yao Wan$^{3}$,
  Jiawan Zhang$^{3}$,
  Kejia Huang$^{5}$,
  Xuming Hu$^{1}$} \\
  $^{1}$Hong Kong University of Science and Technology (Guangzhou)\\
  $^{2}$University of Pennsylvania,
  $^{3}$Huazhong University of Science and Technology\\
  $^{4}$Hong Kong Polytechnic University \\
  $^{5}$Nanjing University of Posts and Telecommunications \\
  \texttt{\{zpling0816, xuminghu97\}@gmail.com}
}

\usepackage{colortbl}

\begin{document}

\maketitle

\begin{abstract}

Large Language Models (LLMs) annotated datasets are widely used nowadays, however, large-scale annotations often show biases in low-quality datasets. For example, Multiple-Choice Questions (MCQs) datasets with one single correct option is common, however, there may be questions attributed to \emph{none} or \emph{multiple} correct options; whereas true-or-false questions are supposed to be labeled with either \emph{True} or \emph{False}, but similarly the text can include unsolvable elements, which should be further labeled as \emph{Unknown}. There are problems when low-quality datasets with mixed question forms can not be identified. We refer to these exceptional label forms as \emph{Sparse Labels}, and LLMs' ability to distinguish datasets with \emph{Sparse Labels} mixture is important. Since users may not know situations of datasets, their instructions can be biased. To study how different instruction settings affect LLMs' identifications of \emph{Sparse Labels} mixture, we introduce the concept of \emph{Instruction Boundary}, which systematically evaluates different instruction settings that lead to biases. We propose \textsc{BiasDetector}, a diagnostic benchmark to systematically evaluate LLMs on datasets with mixed question forms under \emph{Instruction Boundary} settings. Experiments show that users' instructions induce large biases on our benchmark, highlighting the need not only for LLM developers to recognize risks of LLM biased annotation resulting in \emph{Sparse Labels} mixture, but also problems arising from users' instructions to identify them. Code, datasets and detailed implementations are available at \href{https://github.com/ZpLing/Instruction-Boundary}{https://github.com/ZpLing/Instruction-Boundary}.

\end{abstract}

\section{Introduction}

\begin{figure}[!t]
  \centering
  \includegraphics[
      width=\columnwidth,       
      keepaspectratio,
      trim=0 00pt 0 0,         
      clip
  ]{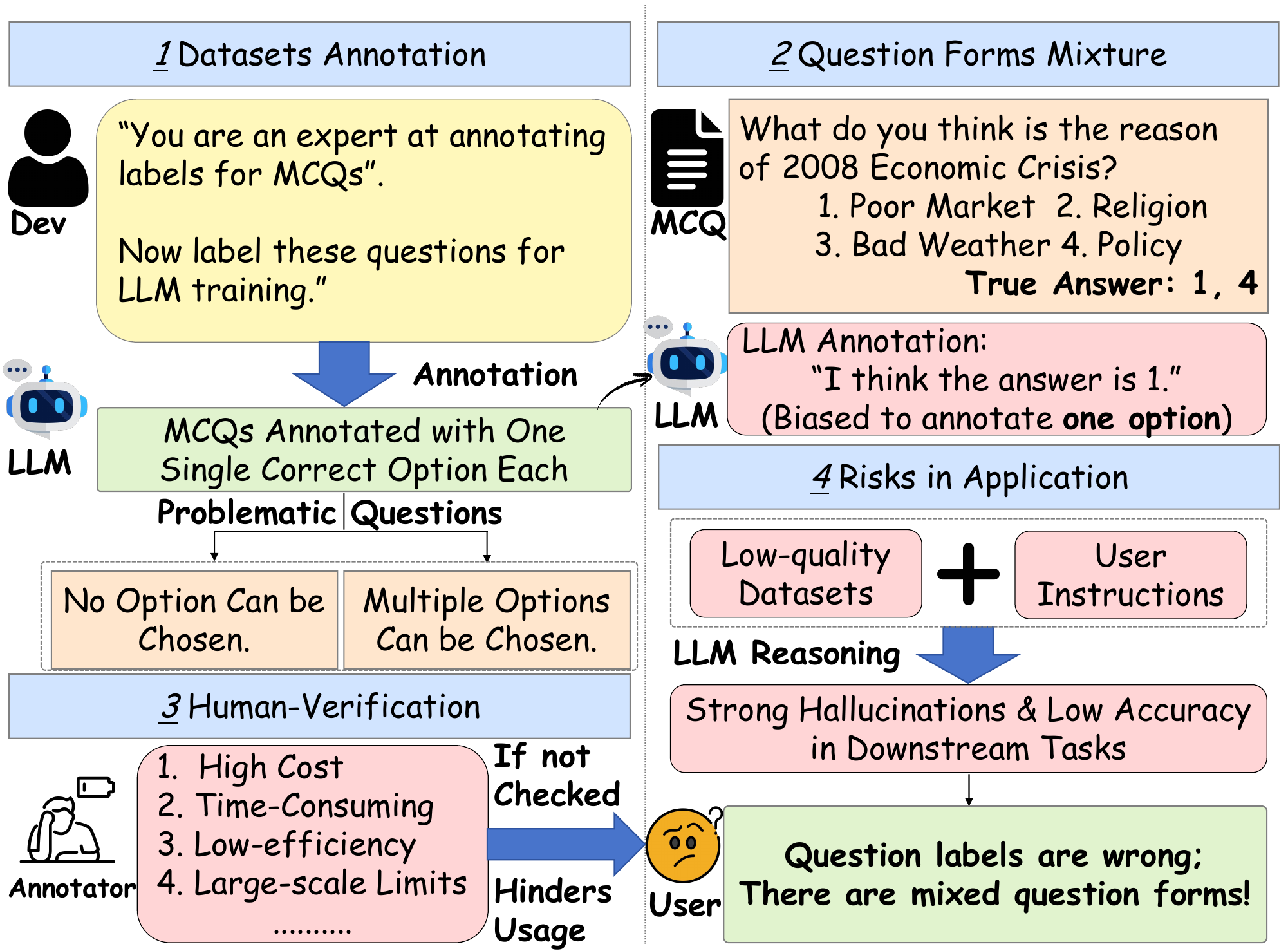}
  \vspace{-10pt} %
  \caption{Background of problems and motivation of our work. Benchmarking the ability of LLMs adaptively answering questions with mixed forms under different instruction settings is important for LLM annotated datasets verification and usage.}
  \label{fig: IB_Formulation}
  \vspace{-10pt}
\end{figure}

LLMs such as GPT-4~\cite{achiam2023gpt} have demonstrated advanced question-answering abilities. However, prior work mainly focuses on thinking along with fixed instruction guidance~\citep{xu2025satabenchselectapplybenchmark,shen-etal-2024-personalized-cloze}: As we tell LLMs they are now facing \emph{MCQs} and choose one option, or \emph{True-or-False} questions which only contains \emph{True} and \emph{False} labels. LLMs are thinking based on a given range, their ability is largely enhanced by such guidance. However, as large-scale LLM annotated datasets are widely used, both text and options may contain flaws, as in Figure~\ref{fig: IB_Formulation}. Low-quality \emph{MCQs} annotated with one single correct option can be common, however, the text and questions may potentially point to \emph{none} or \emph{multiple} correct options. There are contradictions between LLMs instruction following and the complexity of low-quality datasets, which result in strong \emph{Hallucinations} and low accuracy, and such evaluations are lacking.


\begin{figure*}[t]
  \centering
  \includegraphics[
      width=\textwidth,   
      keepaspectratio,
      trim=0 0 0 0,
      clip
  ]{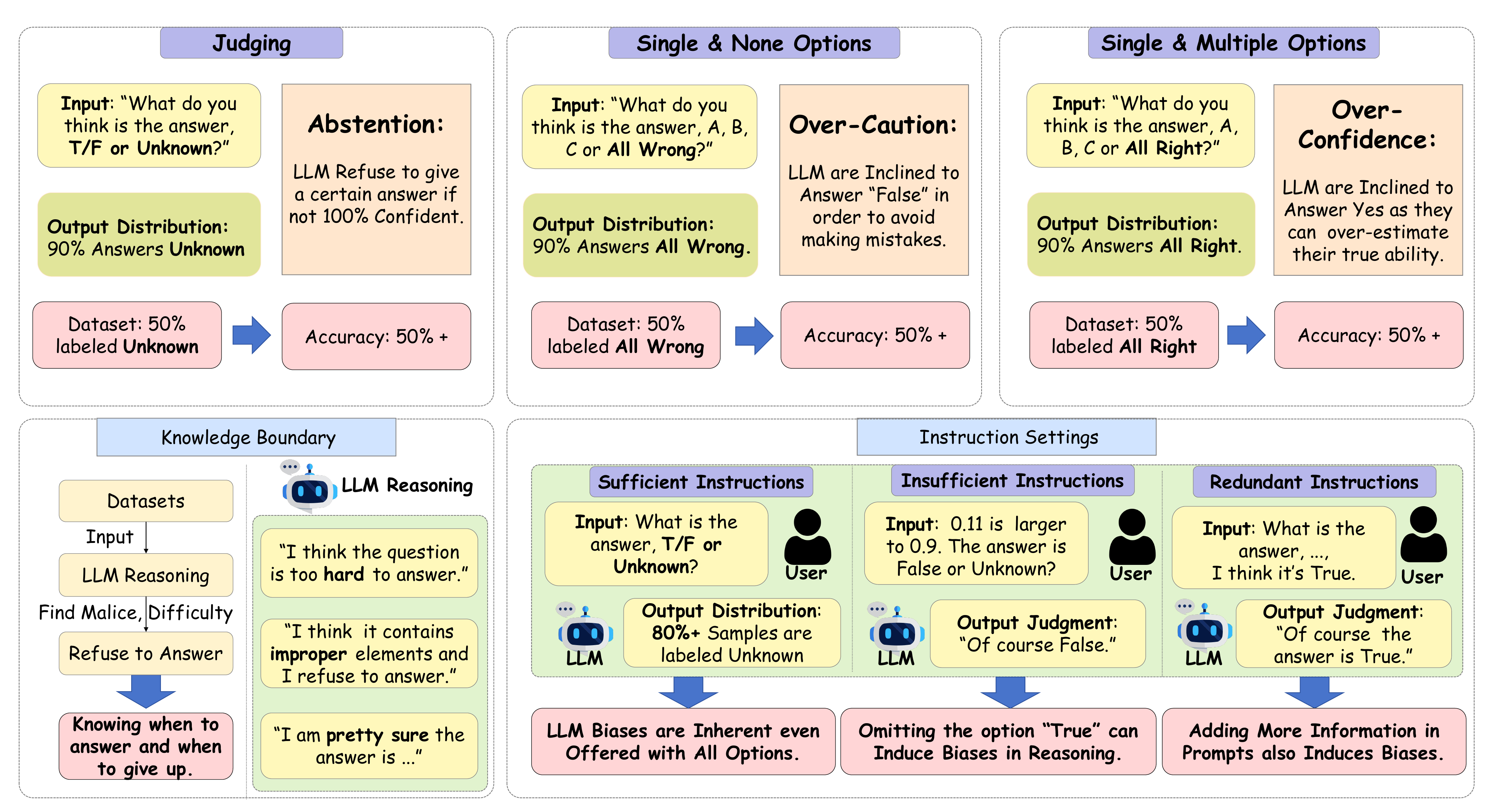}
  \vspace{-10pt}
  \caption{Illustrations of related work concepts. Although LLMs achieve an accuracy of 50\% or higher, it is attributed more to biases than comprehension. \emph{Knowledge Boundary} and \emph{Instruction Settings} indicate LLM biases are influenced by both subjective and objective factors. Detailed Related Work is in Appendix~\ref{related work}.}
  \vspace{-12pt}
  \label{fig:dataset_definition}
\end{figure*}

To address this, we formalize and quantify two factors that influence LLM biases: (1) Datasets with mixed question forms. (2) Different user instruction settings. By simulating the real-world scenarios of users' prompts inducing biases in LLM Reasoning, we establish a benchmark to evaluate LLMs' ability to output correct labels on datasets with mix question forms, under various instruction settings.


As shown in upper three pictures of Figure~\ref{fig:dataset_definition} and~\cite{zheng2024largelanguagemodelsrobust,gupta2023robustllmsincontextmajority}, biased output distributions can lead to seemingly great accuracy (e.g., answering one specific label can achieve 50\% or higher accuracy), however, as output distributions do not align with datasets' label distributions, this only reflects biases and imbalanced accuracy; Prior work on \emph{Abstention}~\citep{wen-etal-2025-know,yin-etal-2023-large}, \emph{Over-Caution}~\citep{cui2025orbenchoverrefusalbenchmarklarge,sullutrone-etal-2025-cover} and \emph{Over-Confidence}~\citep{geng2024surveyconfidenceestimationcalibration} further proves problems arising these concepts. Meanwhile, many papers have explored the notion of the \emph{Knowledge Boundary}~\cite{lewis2020retrieval, zhou2023synthetic}, emphasizing LLMs should acknowledge uncertainty. We study these notions by mixing corresponding questions into our datasets, and study LLMs' performances on our designed datasets under different instruction settings. The latter part is important since users may or may not know there are question forms mixture, thus prompting methods may further induce biases~\citep{li2025humansbrittlelargelanguage}. This evaluation is urgent for both LLM developers and dataset users. 


We formulate usual and exceptional label forms as \emph{Dense Labels} and \emph{Sparse Labels} based on usage statistics in Appendix Table~\ref{Sparse Tense percentage}, and \emph{Instruction Boundary} for studying different instruction settings in Table~\ref{Types of IB}. Compared to the definition of \emph{Knowledge Boundary} on whether LLMs know the exact answers or not, \emph{Instruction Boundary} studying whether instructions cover necessary information matters too. This highlights we consider the phenomenon from both upstream sourcing and downstream processing.

Our work provides a holistic review of how instruction settings can lead to LLM biases on datasets with mixed question forms, which not only analyzes problems regarding LLMs biased annotation and output, but also alerts users to choose proper instruction settings to mitigate biases, pointing to future LLM trustworthiness development. In summary, our contributions are three-fold:

\noindent$\triangleright$ For low-quality datasets and large-scale LLM annotations, we study problems regarding mixed question forms. LLMs are not able to distinguish questions with different label forms, which hinders the usage of LLM-annotated datasets.

\noindent$\triangleright$ For users' instruction settings, we formulate \emph{Instruction Boundary} covering three situations: sufficient, redundant and insufficient instructions. Unlike prior work investigating individual sourcing of prompt or label biases, we propose a unified framework that evaluates LLMs' biases from both user instructions and various question forms mixture.

\noindent$\triangleright$ By comparing \textsc{BiasDetector} with related benchmarks, we conclude how findings across different work generalize, deepening our understandings of LLM trustworthiness. Our work offers values for both LLM developers and dataset users.

\begin{table*}[!t]
  \centering
  \small
  \caption{\emph{Instruction Boundary} settings: classification of instruction settings with descriptions. \emph{Instruction Boundary} is divided into three categories: (1) Sufficient (2) Redundant, and (3) Insufficient Instructions. Each category is further divided into sub-categories. \textbf{Disturbing Miscellany} is carried out separately, based on MMLU-Pro~\cite{wang2024mmluprorobustchallengingmultitask} with 10 options in total.}
    \label{Types of IB}
  \rowcolors{3}{gray!20}{white}
  \begin{tabular}{m{0.18\textwidth} m{0.32\textwidth} m{0.4\textwidth}}
    \toprule
    \midrule
    \addlinespace
    \makecell{\textbf{Sufficient-}\\\textbf{Instructions}} & \makecell{\textbf{Paradigm}} & \makecell{\textbf{Examples}} \\
    \addlinespace
    \midrule
    \toprule
    \addlinespace
    \makecell{\textbf{Vanilla}} &
    Instructions are comprehensive with all label forms and detailed guidance. &
    Which is the answer? True: (Definition...). False: (Definition...). Unknown: (Definition...).\\
    \addlinespace

    \makecell{\textbf{Reasoning-}\\\textbf{Reflection}} &
    Collecting and inputting previous reasoning traces for LLM reflection. &
    Which is the answer? T:..., F:..., or Unknown:... Also show each reasoning step. \textcolor{blue}{(Collects reasoning traces for reflection.)}\\
    \addlinespace

    \makecell{\textbf{Prompt-}\\\textbf{Polishing}} &
    Asking LLMs to polish up prompts to improve understanding. &
    Which is the answer? T:..., F:..., or Unknown:... \textcolor{blue}{(Prompt polished up by LLMs.)}\\ 
    \addlinespace

    \toprule
    \midrule
    \addlinespace
    \makecell{\textbf{Redundant-}\\\textbf{Instructions}} & \makecell{\textbf{Paradigm}} & \makecell{\textbf{Examples}} \\
    \addlinespace
    \midrule
    \toprule
    \addlinespace

    \makecell{\textbf{Conformity}} &
    LLMs exhibit bias toward labels (\emph{dense} or \emph{sparse}) that are explicitly in prompts. &
    Which is the answer? T:..., F:..., or Unknown:... I think is .... \textcolor{blue}{(One \emph{dense} or \emph{sparse label} is directly offered in instructions.)}\\
    \addlinespace

    \makecell{\textbf{Disturbing}\\\textbf{Miscellany}} &
    Additional wrong options are included, which increases distraction. &
    Which is the answer? ..... \textcolor{blue}{(Contains extra wrong options in \emph{MCQs}.)}\\
    \addlinespace

    \makecell{\textbf{Few-shot}\\\textbf{Learning}} &
    LLMs are biased toward labels that are presented in few-shot learning.&
    Which is the answer? T:..., F:..., or Unknown:.... Here is an example to learn from. \textcolor{blue}{(One example with one \emph{dense} or \emph{sparse label}.)}\\
    \addlinespace

    \toprule
    \midrule
    \addlinespace
    \makecell{\textbf{Insufficient-}\\\textbf{Instructions}} & \makecell{\textbf{Paradigm}} & \makecell{\textbf{Examples}} \\
    \addlinespace
    \midrule
    \toprule
    \addlinespace

    \makecell{\textbf{Missing}\\\textbf{Choices}} &
    Some question forms are omitted, causing biases towards available options. &
    Which is the answer? T:..., Unknown:... \textcolor{blue}{(One specific option is omitted.)} \\
    \addlinespace

    \makecell{\textbf{Vagueness}} &
    Instructions cover all question forms but lack detailed guidance. &
    Which is the answer? True, False, or Unknown. \textcolor{blue}{(Detailed guidance is omitted.)}\\
    \addlinespace

    \midrule
    \bottomrule
  \end{tabular}
  \rowcolors{0}{}{}
\end{table*}

\section{Phenomenon Formulation}

\paragraph{Instruction Boundary} \emph{Instruction Boundary} covers three kinds: (1) \textbf{Sufficient Instructions} covers necessary information (i.e., all possible question forms, detailed guidance), where possible reasoning biases mainly arise from the LLMs' internal tendencies. (2) \textbf{Redundant Instructions} with extra and more information, which can potentially misguide LLMs. (3) \textbf{Insufficient Instructions} that leave out specific and necessary information. As shown in Table~\ref{Types of IB}, \emph{Instruction Boundary} covers eight different experimental settings. Details, Templates are in Appendix~\ref{Biases_Details},~\ref{sec:choice_prompts}.

\label{Label Categorization}
\paragraph{Label Categorizations} We classify question label forms into \emph{dense} and \emph{sparse} based on usage statistics in Appendix Table~\ref{Sparse Tense percentage}. \emph{Dense Labels} are often seen and trained on; \emph{Sparse Labels} are infrequently used and treated as exceptional forms. In our work, one correct option in \emph{MCQs} and \emph{True}, \emph{False} labels in \emph{Judging Questions} are \emph{Dense Labels}, whereas none correct option, multiple correct options and the \emph{unknown} labels are categorized into \emph{Sparse Labels}, as shown in Table~\ref{tab:dataset_samples}. Details are in Appendix~\ref{dataset_details}.

\begin{table}[t]
  \centering
  \caption{Datasets categorizations, \emph{Judging Questions} and \emph{MCQs} are tested separately. \emph{Sparse Labels} are in bold.}
  \vspace{-0.25cm}
  \renewcommand{\arraystretch}{1.2}
  \setlength{\tabcolsep}{4pt}
  \small
  \resizebox{\columnwidth}{!}{
    \begin{tabular}{l l l >{\raggedright\arraybackslash}p{3.5cm}}
      \toprule
      \textbf{\large Type} & \textbf{\large Dataset} & \textbf{\large Samples} & \textbf{\large Form} \\
      \midrule

      \multirow{2}{*}{\makecell[l]{\textbf{Judging}\\\textbf{Questions}}}
        & FLD~\cite{morishita2024enhancing} & 600
        & \multirow{2}{*}{\normalsize{\makecell[l]{T/F/\textbf{Unknown}}}} \\
        & FOLIO~\cite{han2022folio} & 640 & \\
      \midrule

      \multirow{6}{*}{\makecell[l]{\textbf{Multiple}\\\textbf{-Choice}\\\textbf{Questions}\\\textbf{(MCQs)}}}
        & ScienceQA~\cite{lu2022learn} & 300
        & \multirow[c]{4}{*}{\normalsize{\makecell[l]{Single-correct}}} \\
        & MMLU~\cite{hendryckstest2021} & 300 & \\
        & TruthfulQA~\cite{lin2021truthfulqa} & 300 & \\
        & MMLU-Pro~\cite{wang2024mmluprorobustchallengingmultitask} & 900 & \\
      \cdashline{2-4}
        & CommonsenseQA~\cite{talmor2019commonsenseqa} & 900
        & \normalsize{\makecell[l]{\textbf{None-correct}}} \\
      \cdashline{2-4}
        & MultiRC~\cite{khashabi2018looking} & 900
        & \normalsize{\makecell[l]{\textbf{Multiple-correct}}} \\
      \bottomrule
    \end{tabular}
  }
  \label{tab:dataset_samples}
  \vspace{-10pt}
\end{table}

\begin{figure*}[!t]
  \centering
  \vspace{-5pt}
  \includegraphics[width=1.0\textwidth]{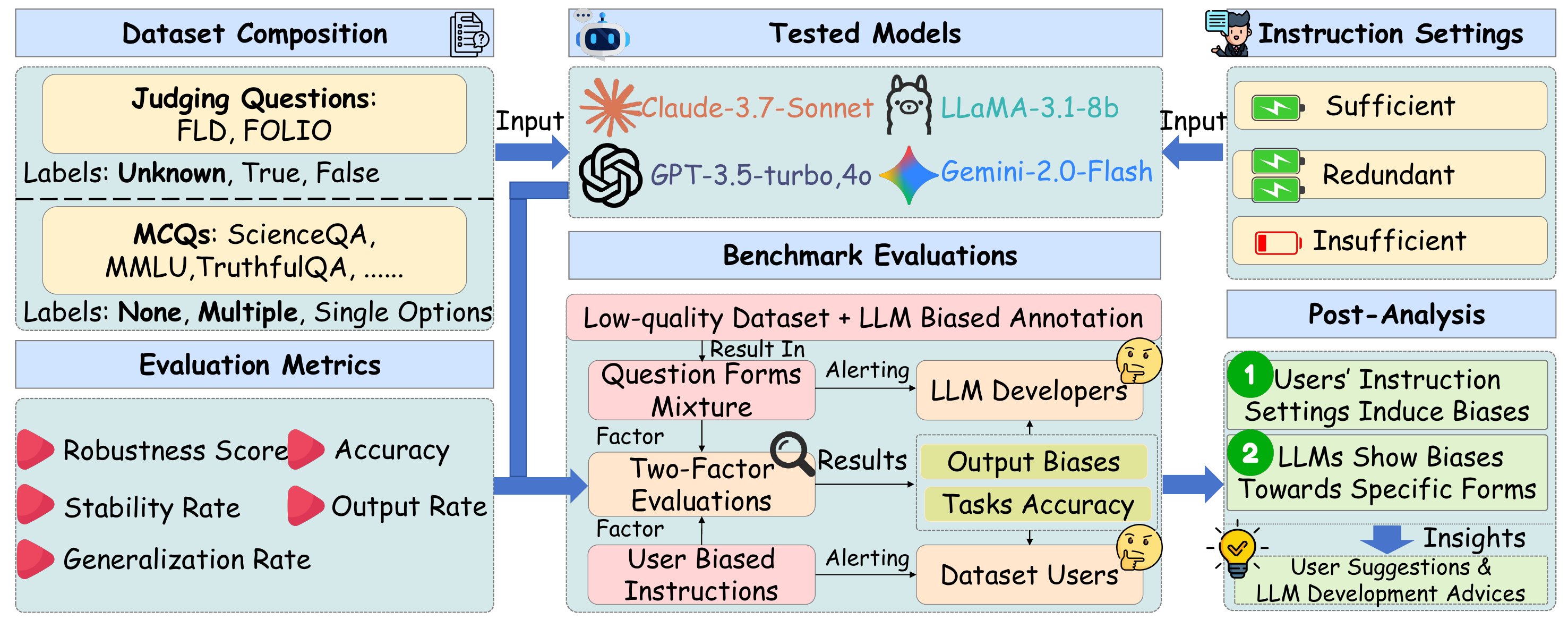}
    \vspace{-15pt}
\caption{General framework of our proposed framework \textsc{BiasDetector}. First of all, we establish datasets with mixed question forms and \emph{Instruction Boundary} settings. Then we apply five LLMs from different families to generate responses. Finally by testing their robustness and accuracy regarding tasks accuracy and output biases, we evaluate problems and offer insights for both LLM developers and dataset users.}
  \label{fig:pipeline}
  \vspace{-10pt}
\end{figure*}

\label{label}
\paragraph{Datasets} Our dataset covers two question types: (1) \emph{Judging Questions}: FLD~\cite{morishita2024enhancing}, and FOLIO~\cite{han2022folio}, contains \emph{True}, \emph{False}, and \emph{Unknown} labels (2) \emph{Multiple-Choice Questions (MCQs)}, containing samples from ScienceQA~\cite{lu2022learn}, MultiRC~\cite{khashabi2018looking}, MMLU~\cite{hendryckstest2021}, MMLU-Pro~\cite{wang2024mmluprorobustchallengingmultitask}, CommonsenseQA~\cite{talmor2019commonsenseqa}, TruthfulQA~\cite{lin2021truthfulqa}. First, we process both types of questions to fit the framework: (1) For \emph{Judging Questions} datasets, these samples are labeled by dataset creators, and we did no revision. (2) For \emph{MCQs}, questions with one correct option are considered to have \emph{Dense Labels}, for none or multiple correct options are categorized as \emph{Sparse Labels}. We extract 300 samples per dataset from ScienceQA, MMLU and TruthfulQA, representing samples with one correct option. We then extract 900 samples from CommonsenseQA containing five options with one correct option, by deleting the correct one we get samples with none correct option. For multiple correct options, we extract 900 samples from MultiRC with four options with two correct ones. Sampling seeds are all set to 42, revised samples are manually checked to ensure validness, examples are in Appendix~\ref{Post-Processed Dataset Examples}. Details are in Table~\ref{tab:dataset_samples}. \textbf{As we simulate question forms mixture, adding options like "None Correct" or "A and B are Correct"  does not meet real-world scenarios. LLMs should identify different forms based on the same instruction for each question, rather than being informed separately in options.}

\begin{table*}[!t]
  \centering
  \caption{Results across seven settings except \emph{Disturbing Miscellany}. Accuracy is presented with the Standard Deviation (std) by running each setting four times repeatedly to mitigate randomness. Highest values are in bold.}
  \label{tab:main results}
  \vspace{-7pt}
  \small
  \resizebox{\textwidth}{!}{%
  \begin{tabular}{@{}l l c c c c c c c@{}}
    \toprule
    \multirow{2}{*}{\textbf{Models}} & 
    \multirow{2}{*}{\textbf{Settings}} &
    \multicolumn{3}{c}{\textbf{Judging Questions}} &
    \multicolumn{4}{c}{\textbf{MCQs}} \\
    \cmidrule(lr){3-5}\cmidrule(lr){6-9}
        & & \textbf{$Acc$} & \textbf{$SR$} & \textbf{$GR$} &
            \textbf{$Acc$} & \textbf{$SR$} & \textbf{$GR^1$} & \textbf{$GR^2$} \\
    \midrule
\rowcolor{gray!20}
\multirow{9}{*}{\makecell{\textbf{GPT-3.5-turbo}}}
& Vanilla          & $48.75 \pm 2.5$ & 21.47 & 76.03 & {$24.67 \pm 1.3$} & 62.00 & 12.00 & 0.00 \\ 
\rowcolor{gray!20}
& Reasoning-Reflection  & $36.14 \pm 1.3 $ & 31.51 & 40.78 & {$33.56 \pm 1.5$} & 40.67 & 60.00 & 0.00 \\
\rowcolor{gray!20}
& Prompt-Polishing & $34.30 \pm 1.3 $ &  9.16 & 59.44 & {$21.33 \pm 1.3$} & 58.67 &  5.33 & 0.00 \\ 
\cdashline{2-9}
\rowcolor{gray!20}
& Conformity-Dense       & $50.00 \pm 0.0$ &  9.38 & 90.63 & {$32.20 \pm 1.8$} & 83.33 & 13.33 & 0.00 \\ 
& Conformity-Sparse       & $50.47 \pm 1.2$ &  1.56 & \textbf{99.38} & {$37.11 \pm 1.8$} & 11.33 & \textbf{100.00} & 0.00 \\ 
\rowcolor{gray!20}
& Few shot-Dense& $50.00 \pm 0.0$  & \textbf{99.38} &  0.63 & {$35.60 \pm 2.0$} & 93.33 & 16.67 & 0.00 \\ 
& Few shot-Sparse& $50.47  \pm 0.0$  & 1.56 &  \textbf{99.38} & {$52.25 \pm 2.0$} & 90.00 & 43.35 & 0.00 \\ 
\cdashline{2-9}
\rowcolor{gray!20}
& Missing Choices  & $34.76 \pm 1.3$ & 61.04 &  8.48 & {$25.56 \pm 2.8$} & 64.00 & 12.67 & 0.00 \\ 
\rowcolor{gray!20}
& Vagueness        & $38.32 \pm 0.8$ & 23.17 & 50.38 & {$24.70 \pm 2.2$} & 73.30 &  0.70 & 0.00 \\ 
\midrule
\multirow{9}{*}{\makecell{\textbf{GPT-4o}}}
& Vanilla          & $45.41 \pm 2.1$ & 63.49 & 27.32 & {$35.60 \pm 2.8$} & 89.33 & 28.67 & 0.00 \\ 
& Reasoning-Reflection  & $40.07 \pm 2.4$ & 29.53 & 50.62 & {$18.89 \pm 2.0$} & 16.67 & 40.00 & 0.00 \\ 
& Prompt-Polishing & $48.86 \pm 1.9$ & 60.50 & 37.21 & {$37.11 \pm 2.0$} & 89.33 & 19.33 & 2.67 \\ 
\cdashline{2-9}
& Conformity-Dense       & $55.31 \pm 1.9$ & 62.19 & 48.44 & {$40.00 \pm 1.5$} & 90.00 & 30.00 & 0.00 \\ 
& Conformity-Sparse       & $56.92 \pm 1.9$ & 20.60 & 93.24 & {$44.00 \pm 1.5$} & 83.33 & 48.67 & 0.00 \\ 
& Few shot-Dense & $56.12 \pm 1.8$ & 80.73 & 31.50 & {$35.60 \pm 1.5$} & 93.30 & 13.30 & 0.00 \\ 
& Few shot-Sparse & $63.13 \pm 1.8$ & 61.25 & 65.00 & {$44.00 \pm 1.5$} & \textbf{98.00} & 34.00 & 0.00 \\ 
\cdashline{2-9}
& Missing Choices  & $40.39 \pm 2.4$ & 38.09 & 47.69 & {$40.89 \pm 1.5$} & 90.00 & 32.67 & 0.00 \\ 
& Vagueness        & $41.01 \pm 2.2$ & 12.28 & 69.75 & {$32.70 \pm 2.8$} & 93.30 &  4.70 & 0.00 \\ 
\midrule
\rowcolor{gray!20}
\multirow{9}{*}{\makecell{\textbf{LLaMA-3.1-8B}}}
& Vanilla          & $51.87 \pm 2.2$ & 17.92 & 85.82 & {$20.67 \pm 2.2$} & 55.33 &  4.67 & 2.00 \\ 
\rowcolor{gray!20}
& Reasoning-Reflection  & $38.81 \pm 1.5$ & 57.70 & 19.93 & {$15.33 \pm 1.5$} & 36.00 &  8.67 & 1.33 \\ 
\rowcolor{gray!20}
& Prompt-Polishing & $42.58 \pm 0.8$ & 50.28 & 34.88 & {$7.78 \pm 0.0$} & 13.33 &  5.33 & 4.67 \\ 
\cdashline{2-9}
\rowcolor{gray!20}
& Conformity-Dense       & $30.47 \pm 1.4$ &  60.31 & 0.63 & {$23.30 \pm 2.8$} & 56.70 & 13.30 & 0.00 \\
& Conformity-Sparse       & $50.00 \pm 0.0$ &  0.00 & 100.00 & {$38.00 \pm 2.8$} & 50.67 & 60.00 & 3.33 \\ 
\rowcolor{gray!20}
& Few shot-Dense & $46.28 \pm 1.2$ & 77.40 & 15.15 & {$26.22 \pm 2.8$} & 70.67 & 6.00 & 2.00 \\ 
& Few shot-Sparse & $51.88 \pm 1.2$ & 5.63 & 98.13 & {$38.00 \pm 2.8$} & 50.67 & 63.33 & 0.67 \\ 
\cdashline{2-9}
\rowcolor{gray!20}
& Missing Choices  & $47.91 \pm 1.5$ &  8.67 & 86.67 & {$22.44 \pm 0.8$} & 59.33 &  5.33 & 2.67 \\ 
\rowcolor{gray!20}
& Vagueness        & $29.21 \pm 2.5$ &  0.32 & 58.11 & {$25.60 \pm 1.5$} & 75.30 &  0.70 & 0.70 \\ 
\midrule
\multirow{9}{*}{\makecell{\textbf{Claude-3.7-}\\\textbf{Sonnet}}}
& Vanilla          & $46.23 \pm 1.8$ & 40.13 & 52.33 & {$7.78 \pm 0.0$} & 12.67 &  7.33 & 3.33 \\ 
& Reasoning-Reflection  & $45.73 \pm 1.3$ & 55.48 & 35.98 & {$14.00 \pm 1.5$} & 16.67 & 24.67 & 0.67 \\ 
& Prompt-Polishing & \textbf{57.21} $\pm$ \textbf{0.8} & 84.06 & 27.66 & {$2.67 \pm 0.0$} &  1.33 &  5.33 & 1.33 \\ 
\cdashline{2-9}
& Conformity-Dense     & $57.03 \pm 1.2$ & 33.75 & 80.31 & {$10.00 \pm 3.7$} &  13.33 &  10.00 & \textbf{6.67} \\ 
& Conformity-Sparse      & $55.10 \pm 1.2$ & 18.70 & 91.49 & {$3.78 \pm 3.7$} &  1.33 &  7.33 & 2.67 \\ 
& Few shot-Dense& $54.10 \pm 1.0$ & 54.09 & 54.11 & {$25.11 \pm 1.8$} & 52.67 & 21.33 & 1.33 \\ 
& Few shot-Sparse& $49.84 \pm 2.5$ & 63.44 & 36.25 & {$22.20 \pm 1.8$} & 55.00 & 5.00 & 6.65 \\ 
\cdashline{2-9}
& Missing Choices  & $53.19 \pm 1.5$ & 26.32 & 80.07 & {$9.78 \pm 1.8$} & 14.67 & 11.33 & 3.33 \\ 
& Vagueness        & $51.20 \pm 0.9$ & 45.34 & 57.07 & {$30.20 \pm 2.5$} & 88.70 &  2.00 & 0.00 \\ 
\midrule
\rowcolor{gray!20}
\multirow{9}{*}{\makecell{\textbf{Gemini-2.0-}\\\textbf{Flash}}}
& Vanilla          & $51.81 \pm 1.8$ & 36.99 & 66.62 & {$35.00 \pm 1.8$} & 84.85 & 18.18 & 2.94 \\ 
\rowcolor{gray!20}
& Reasoning-Reflection  & $40.84 \pm 1.6$ & 39.81 & 41.87 & {|c}{$36.00 \pm 2.5$} & 60.67 & 45.33 & 2.00 \\ 
\rowcolor{gray!20}
& Prompt-Polishing & $35.40 \pm 1.6$ &  0.50 & 70.31 & {$10.44 \pm 0.0$} & 18.67 &  7.33 & 5.33 \\ 
\cdashline{2-9}
\rowcolor{gray!20}
& Conformity-Dense      & $35.00\pm 2.2$ &  56.25 & 13.75 & {$48.90 \pm 1.8$} & 90.00 & 56.70 & 0.00 \\ 
& Conformity-Sparse       & $51.20 \pm 2.2$ &  4.68 & 97.70 & {\textbf{50.00} $\pm$ \textbf{1.8}} & 65.33 & 84.67 & 0.00 \\ 
\rowcolor{gray!20}
& Few shot-Dense& $54.51 \pm 2.5$ & 88.23 & 18.31 & {$42.67 \pm 1.5$} & 87.33 & 40.67 & 0.00 \\ 
& Few shot-Sparse& $57.34 \pm 1.5$ & 17.19 & 78.13 & {$42.67 \pm 1.5$} & 87.33 & 40.67 & 0.00 \\ 
\cdashline{2-9}
\rowcolor{gray!20}
& Missing Choices  & $45.79 \pm 0.8$ & 31.30 & 60.28 & {$36.00 \pm 2.8$} & 87.88 & 21.21 & 0.00 \\ 
\rowcolor{gray!20}
& Vagueness        & $40.73 \pm 1.5$ & 19.49 & 61.96 & {$31.60 \pm 1.8$} & 90.70 &  4.00 & 0.00 \\ 
\bottomrule
\end{tabular}%
}
\vspace{-5pt}
\end{table*}

\section{Benchmark Metrics}
\label{Metrics}
We introduce the metric of performance \textbf{Accuracy}:
\begin{equation}
Acc = \frac{1}{|D|} \sum_{i=1}^{|D|} \mathbbm{1} \!\left(y^{i} = \hat{y}^{i}\right)
\end{equation}
Further, we implement the following two metrics, defining \textbf{Stability Rate (SR)} and \textbf{Generalization Rate (GR)} as below, measuring the accuracy of two label forms respectively:
\begin{equation}
SR = \frac{1}{|D_D|} \sum_{y^{i} \in D_{D}} \mathbbm{1}\!\left(y^{i} = \hat{y}^{i}\right)
\end{equation}
\begin{equation}
GR = \frac{1}{|D_S|} \sum_{y^{i} \in D_{S}} \mathbbm{1}\!\left(y^{i} = \hat{y}^{i}\right)
\end{equation}
For y$^{i}$ indicates the true label, $\hat{y}^{i}$ is the predicted label, and $D_{D}$, $D_{S}$ mean datasets with \emph{dense} and \emph{sparse} labels, which are 50\% and 50\% equally sampled, and \emph{D} indicates the whole dataset, from which we can get $D_{\mathrm{D}} \cup D_{\mathrm{S}} = D$. \emph{GR} represents LLMs identify \emph{Sparse Labels} correctly, and \emph{SR} for \emph{Dense Labels}. Since there are two \emph{sparse label} forms in \emph{MCQs}, we further decompose its \emph{GR} into $GR^1$ and $GR^2$, measuring None Correct Option and Multiple Correct Options.
Beside, we study biases via output distributions \textbf{Output Rate (OR)}. We analyze \textbf{output label distributions} of \emph{Judging Questions} and \textbf{output option number distributions} of \emph{MCQs}:
\begin{equation}
OR_D = \frac{1}{|D_D|}\sum_{i=1}^{|D_D|} \mathbbm{1}\!\left(\hat{y}^{i} \in D_D\right)
\end{equation}
\begin{equation}
OR_S = \frac{1}{|D_S|}\sum_{i=1}^{|D_S|} \mathbbm{1}\!\left(\hat{y}^{i} \in D_S\right)
\end{equation}
$OR_D$ and $OR_S$ represents LLM output distributions of \emph{Dense Labels} and \emph{Sparse Labels}. Outputting  \emph{one} option in \emph{MCQs}, or \emph{True} and \emph{False} labels in \emph{Judging Questions} are categorized into $OR_D$, and other output situations fall into $OR_S$.


\section{Experiments}

\subsection{Experimental Setup}

\paragraph{Models}
As in~\cite{ling2025wakenllmevaluatingreasoningpotential}, although larger parameter LLMs show better performances, they are willing to output \emph{unknown}. We test closed and open-source LLM families, including Claude-3-7-Sonnet~\cite{anthropic2024claude3}, Gemini-2.0-Flash~\cite{google2024gemini2flash}, LLaMA-3.1-8B~\cite{touvron2023llama}, GPT-3.5-turbo and GPT-4o~\cite{openai2023gpt35, openai2025gpt4o}. Tested models use Temperature = 0.5 unless specified; Qwen-2.5-7b~\cite{bai2023qwentechnicalreport} used as a LLM judge runs with T=0, and we apply testing cases to prove its validness. Since the LLM judge deals with exceptional cases when \textbf{label names} in \emph{Judging Questions} or \textbf{option indexes} in \emph{MCQs} failed to match, the usage is small (below 5\%) and biases are minor. Details are in Appendix~\ref{LLM Judge Verifications}.

\paragraph{Experimental Settings} As in Table~\ref{Types of IB}, we test LLMs under \emph{Instruction Boundary} settings to quantify biases attributed to both instructions and datasets with mix question forms, except for the \emph{Vanilla} setting served as a baseline. \emph{Judging Questions} and \emph{MCQs} are tested separately to better simulate real-world scenarios. Case studies are in Appendix~\ref{Output Cases}.

\subsection{General Experimental Results}
As shown in Figure~\ref{tab:main results}, tested LLMs show poor performance as the accuracy lies in the range of $35\%$ to $60\%$. Claude-3.7-sonnet, GPT-4o shows the highest average $Acc$. Meanwhile, LLMs show high instability and reasoning gap between \emph{dense} and \emph{sparse} labels since the gap between $SR$ and $GR$ is high. The value of $GR^{2}$, which is the accuracy in multiple correct options, are mostly less than 5\%, indicating a strong \textbf{\emph{Over-Caution}} behavior.

For \emph{Judging Questions}, the $SR$ is generally lower than $GR$, indicating given the \emph{Unknown} option, LLMs are inclined to this rather than outputting \emph{True} or \emph{False}. This situation is slightly better for GPT-4o and Claude-3.7-Sonnet, as the gap between two metrics is relatively low. We also observe instability in GPT-3.5-turbo, as $SR$ and $GR$ values are completely different in \emph{Conformity} and \emph{Few-shot Learning}, showing strong biases if being misguided.

Reasoning on \emph{MCQs} demonstrates an opposite behavior, as $SR$ is overall higher than $GR^1$ and $GR^2$, LLMs are good at answering questions with one correct option over others, and this gap is extremely huge for GPT-4o and Gemini-2.0-Flash. We also observe a trade-off for these models: Being better at reasoning \emph{Judging Questions} means worse stability facing \emph{MCQs}.
\paragraph{Significance Testing} By comparing the results of the \emph{Vanilla} setting with other biased settings, we gain better understanding of how instruction settings influence LLMs. As in Figure~\ref{fig:bar results}, some cause serious accuracy drop, and others result in a seemingly good accuracy with a biases shifting. Overall, we can see \emph{Vagueness}, \emph{Reasoning-Reflection} and \emph{Disturbing Miscellany} influence accuracy the most.

\subsection{Results on Sufficient Instructions}
\begin{figure*}[t]
  \centering

  \includegraphics[width=1.0\textwidth]{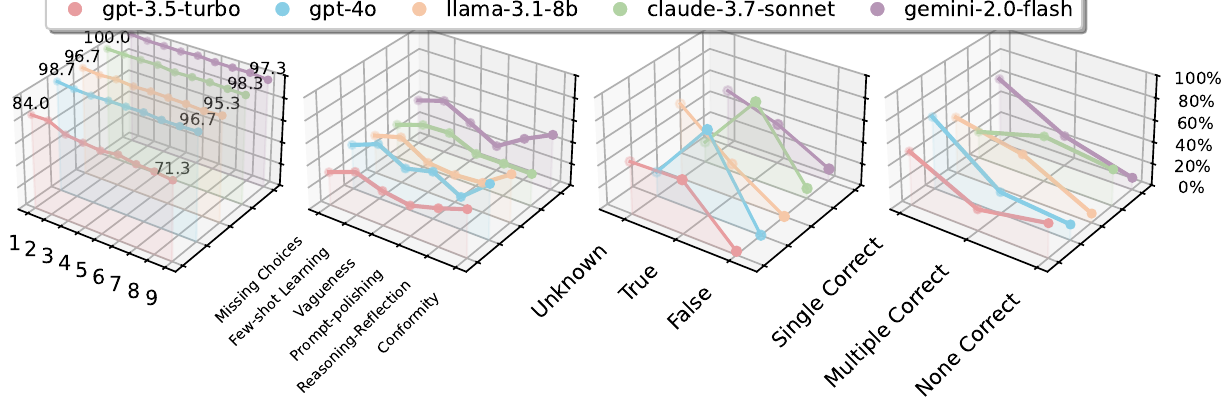}
  \vspace{-20pt}
  \caption{Results of \emph{Accuracy} under \emph{Disturbing Miscellany} setting, accuracy is dropping as option number is increasing according to the x-axis, and average \emph{Accuracy} among settings (left one, two). \emph{Label Output Distribution (OR)} under the \emph{Vanilla} setting (right one, two). Standardized with the scale of the rightmost figure as the reference.}
  \label{fig:side results}
  
  \includegraphics[width=1.0\textwidth]{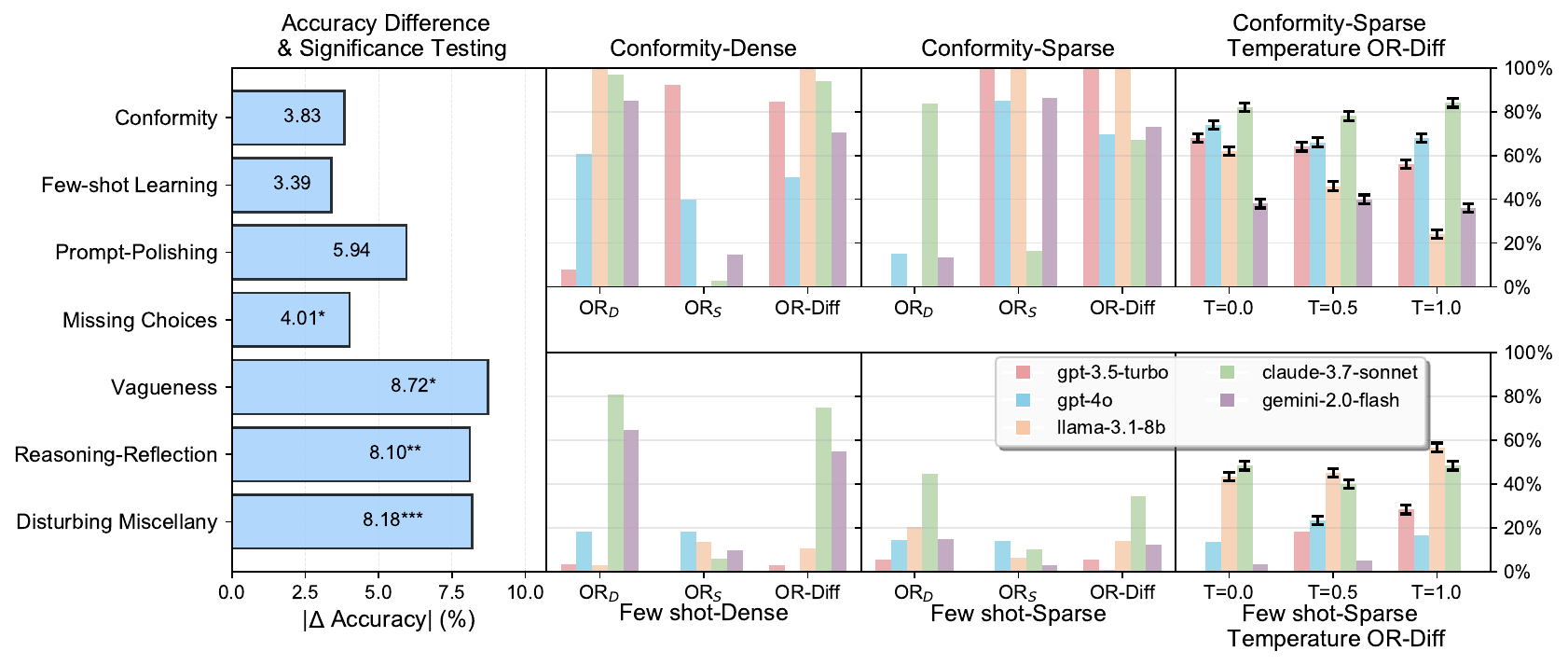}
  \vspace{-15pt}
  \caption{Accuracy differences comparing biased settings with the \emph{Vanilla} setting, and Significance Testing. *: p < 0.05 (95\% confidence); **: p < 0.01 (99\% confidence), ***: p < 0.001 (99.9\% confidence). Detailed results on \emph{MCQs} under \emph{Conformity} and \emph{Few-shot} settings (Right Six). $OR-Diff =$  |$OR_D$ - $OR_S$|. Details are in Section~\ref{Metrics}. }
  \label{fig:bar results}

  \vspace{-10pt}

\end{figure*}

Prior work~\citep{zheng2024largelanguagemodelsrobust,fei2023mitigatinglabelbiasesincontext} demonstrated LLMs show biases towards specific labels, and self-reflection~\citep{xu2024prideprejudicellmamplifies}, prompt polishing can induce biases and instability~\citep{Jakesch_2023}. But LLMs' ability of distinguishing question forms mixture are rarely tested.
\label{Vanilla Scenario}
\paragraph{Vanilla Setting} As illustrated in the right two panels of Figure~\ref{fig:side results}, LLMs exhibit explicit and unified bias even when instructions cover all possible question forms: for \emph{Judging Questions}, LLMs are inclined to output \emph{Unknown} and \emph{True}. For \emph{MCQs}, outputting one single choice takes up a significant amount, indicating strong and unified biases towards specific labels, and simply including all necessary elements for LLMs to choose from is not enough for mitigating biases.


\paragraph{Reasoning-Reflection} As in Table~\ref{tab:main results}, while this setting can narrow the gap between \emph{SR} and \emph{GR}, it yields mostly accuracy drop, indicating LLM reflecting on previous reasoning traces is problematic.

\paragraph{Prompt-Polishing} Users like to polish up prompts by LLMs. We input prompts in \emph{Vanilla} setting into GPT-4o and ask LLMs to polish up them, templates are in Appendix~\ref{LLM Judge Templates}. We observe a wide accuracy drop among all models, ranging from 5\% to 30\%. Meanwhile, most LLMs demonstrate a larger gap between \emph{SR} and \emph{GR} compared with the \emph{Vanilla} setting, indicating this setting is likely to induce biases.

Even with complete instruction coverage, LLMs continue to exhibit consistent biases, suggesting the performances on mixed question forms is weak, and the ability to adaptively reason needs improvements.

\subsection{Results on Redundant Instructions}
\label{Redundant Instructions}
LLMs show biases towards few-shot examples was addressed in~\citep{zhao2021calibrateuseimprovingfewshot,reif2024performancequantifyingmitigatinglabel,si2023measuringinductivebiasesincontext}, meanwhile, conformity effects are researched in~\citep{zhu2024conformity, shoval2025conformityTrial,sharma2025understandingsycophancylanguagemodels}, and robustness on option numbers misguidance is in~\citep{zhang2024multiplechoicequestionsefficientrobust}. However, biases facing mixed question forms is still under-explored. To the best of our knowledge, we are the first to compare \emph{Few-shot Learning} with \emph{Conformity} and extra options in \emph{MCQs} to study biases, and we further sub-categorize \emph{Few-shot Learning} and \emph{Conformity} based on included label forms.

\paragraph{Few-shot Learning} As in Table~\ref{tab:main results}, compared to the \emph{Vanilla} setting, Few-shot learning leads to an accuracy gain, but the value of \emph{SR} and \emph{GR} are largely changed. Details are in Appendix~\ref{Few-shot details}. We include examples with either \emph{sparse} or \emph{dense} labels and compare differences, Figure~\ref{fig:bar results} shows that example labels can trigger huge biased output. LLM performance gain is largely attributed to bias shifts rather than understanding. The ability cannot be improved via imitations since performances are unstable.

\paragraph{Conformity} LLM biases can be extremely high when labels are given directly. Figure~\ref{tab:main results} indicates that even though accuracy doesn't change much, the gap between \emph{SR} and \emph{GR} is even higher compared with other settings. Details of \emph{OR} are in Fig~\ref{fig:bar results}, charts 2 and 3 demonstrate a higher \emph{OR} shift in most LLMs regarding both \emph{dense} and \emph{sparse} labels. LLMs demonstrate much higher conformity regarding \emph{Sparse Labels} like \emph{Unknown}, compared with \emph{Dense Labels}. We also quantify how \emph{(OR)} varies under three temperature settings: $T \in \{0.0,\,0.5,\,1.0\}$. As in Figure~\ref{fig:bar results}, takeaways regarding temperature changes is same for both \emph{Few-shot} and \emph{Conformity}: merely revising $T$ yields limited gains, and larger temperatures tend to introduce greater instability.

\begin{figure*}[!t]
  \centering
  \includegraphics[width=1.0\textwidth]{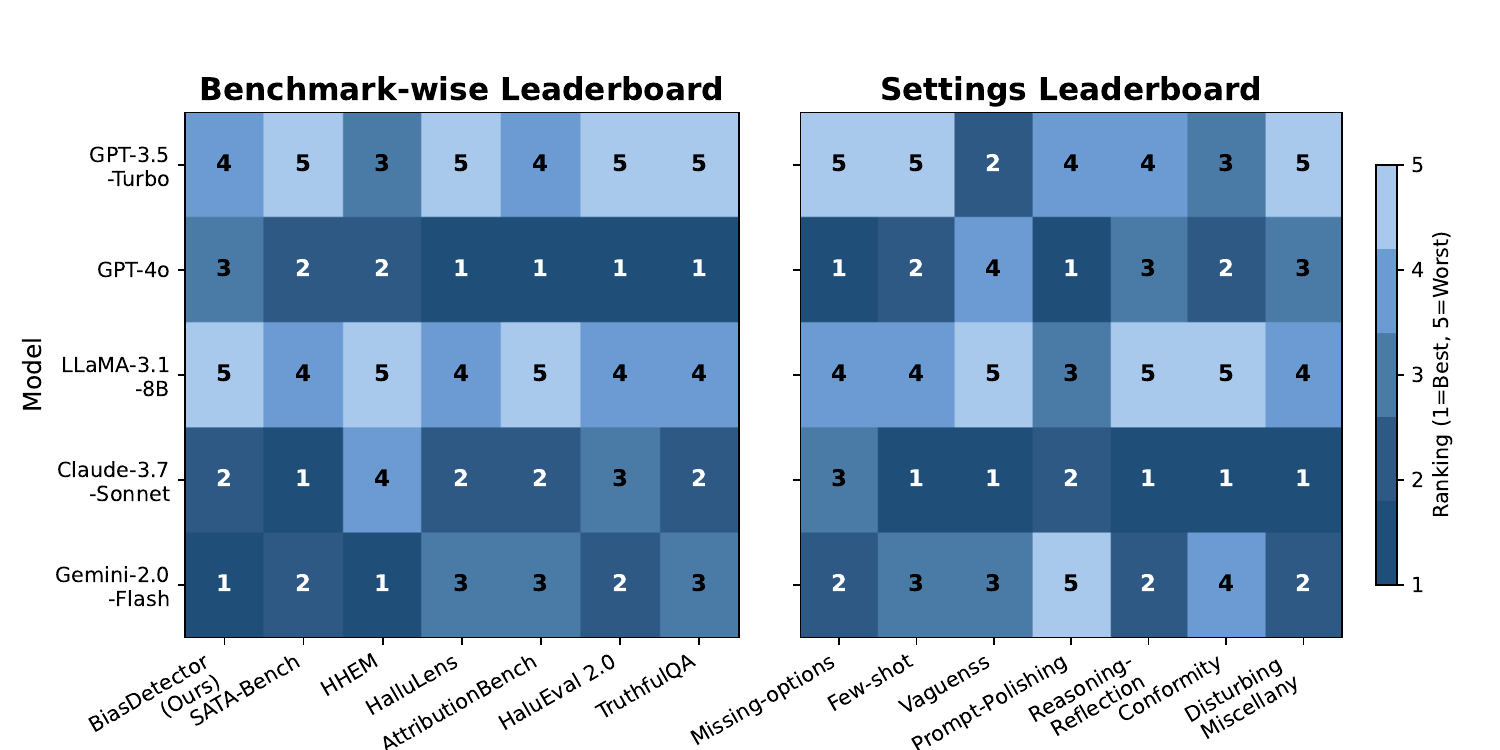}
  \vspace{-15pt}
  \caption{Leaderboard of related topic benchmarks(left), and inner settings comparisons among five LLMs used in our benchmark(Right). For benchmark-wise comparisons, we show the ranking under the \textbf{Vanilla Setting}. Our metric is \textbf{Robustness Score (RS)}. Detailed values can be found in Table~\ref{tab:detailed-rs}.}
  \label{fig:leaderboard}

  \vspace{-10pt}

\end{figure*}

\paragraph{Disturbing Miscellany} We investigate how the number of wrong options (i.e., information redundancy) affects accuracy. We choose MMLU-Pro~\cite{wang2024mmluprorobustchallengingmultitask}, \emph{MCQs} consisting one true option and nine false options per sample. We save \textbf{N} numbers of wrong options to quantify how redundant information influences performances, \textbf{N} ranges from one to nine. As in Figure~\ref{fig:side results}, most LLMs show a slight drop in accuracy, except for outdated models like GPT-3.5-turbo demonstrating a high decline, others maintain high robustness when \emph{N} increases.
This setting proves LLMs' resistance to redundancy is strong. As \textbf{\emph{Conformity} induces the most serious biases, \emph{Few-shot Learning} ranks the second, and \emph{Disturbing Miscellany} the last.} We empirically show that \emph{Conformity} effects result from expression tuning rather than redundant information.

\subsection{Results on Insufficient Instructions}
Prior research on prompt vagueness~\citep{si-etal-2023-measuring} and missing options~\citep{góral2025waitthatsoptionllms,cao2024worstpromptperformancelarge} show that LLM performances are largely influenced by prompts, but mixed question forms are rarely tested. This setting is for simulating users not knowing question forms mixture in datasets, and test whether LLMs can independently reason under biased instruction settings. For \emph{Vagueness} and \emph{Missing Choices}, omitting either guidance or \emph{Sparse Label} options result in huge accuracy drop. The primary bottleneck is the ability to think independently. Ideally, even if specific options are omitted in prompts, LLMs can still answer correctly.



\section{Leaderboard Analysis}
 We compare LLM rankings among related benchmarks. Our metric is the \emph{harmonic mean} of \emph{GR} and \emph{SR}, denoted as \textbf{Robustness Score (RS)}:
\begin{equation}
    RS_M=\frac{3\cdot SR \cdot GR^1 \cdot GR^2}{SR\cdot GR^1 + SR\cdot GR^2 + GR^1 \cdot GR^2}
\end{equation}
\begin{equation}
RS_J =  \frac{2\cdot SR \cdot GR}{SR + GR}, RS = \frac{1}{2} (RS_M + RS_J)
\end{equation}
\emph{Judging Questions} and \emph{MCQs} are calculated separately then average. Table~\ref{tab:dataset_samples} shows \emph{GR} and \emph{SR} are calculated based on equal numbers of samples.

We select open-sourced benchmarks (1) Measured by Accuracy: \citep{li2023haluevallargescalehallucinationevaluation}, \citep{bang2025hallulensllmhallucinationbenchmark},~\citep{lin2021truthfulqa}, ~\citep{hhem-2.1-open},~\citep{xu2025satabenchselectapplybenchmark} (2) Measured by Macro-F1: \citep{li2024attributionbenchhardautomaticattribution}. As in Figure~\ref{fig:leaderboard}, our findings mostly align with prior work. For the \emph{Vanilla} setting, Gemini-2.0-Flash offers the greatest cost-efficiency, as the performance ranks first with lower costs compared to GPT-4o and Claude-3.7-Sonnet. 

\section{Discussion}
We offer three advice for problem mitigation: (1) We design a prompt template and run experiments in Appendix~\ref{Proposed Template} to ease biases. (2) LLMs perform poorly on mixed question forms, making datasets manual verifications necessary. (3) MoE (Mixture-of-Experts) structures~\citep{mu2025comprehensivesurveymixtureofexpertsalgorithms} which are suitable for various but interrelated tasks is a promising way to fundamentally solve this problem.

\section{Conclusion}
We establish a dataset containing both \emph{Judging Questions} and \emph{MCQs} with mixed question forms. By evaluating LLMs' ability to adaptively answer correct labels under three instruction settings: sufficient, redundant and insufficient, we systematically quantify LLM weaknesses as they neither think independently under different instruction settings, nor adaptively distinguish mixed question forms. Unlike prior work, we do not revise options to include these forms, instead we test LLMs based on unified instructions to identify forms mixture themselves. Sufficient instructions still introduce biases despite detailed guidance, whereas redundant and insufficient instructions amplify biases, proving the necessity of manual verification for low-quality datasets, and alerts users of instruction settings selection. To the best of our knowledge, we are the first to systematically evaluate LLMs under this setting, offering value for LLM trustworthiness developments.

\section{Limitations}
Although we address three categories of prompt coverage problems, our work does not consider mainstream workflows such as LLM agents or RAG systems. Moreover, our study focuses primarily on language-related tasks, leaving vision-related challenges outside the scope of this paper. Besides, for real-life applications, user prompt forms may be varied and noisy, leaving the final performance instable. Unified analysis regarding different elaboration forms is valuable but too large to be examined.

\section{Ethical Considerations}
All datasets are \textbf{publicly available for non-commercial research}. FLD is an entirely synthetic dataset; FOLIO, ScienceQA, MMLU, CommonsenseQA, TruthfulQA, MMLU-Pro, and MultiRC derive from publicly available open educational resources and contain no personally identifiable information (PII). No additional human data were collected, so the study does not require Institutional Review Board (IRB) review.

\bibliography{custom}

\appendix
\section{Dataset Details}
\label{dataset_details}

\subsection{Overview}

The benchmark is carried out on the following four publicly available reasoning datasets:

\begin{itemize}[leftmargin=0pt,nosep]

  \item \textbf{FLD}~\citet{morishita2024enhancing} is a synthetic \textbf{Fact-based
        Logical Deduction} corpus comprising multi-step deductive problems derived from first-order predicate
        logic axioms. Each instance is labeled as \textbf{True},
        \textbf{False}, or \textbf{Unknown}, enabling fine-grained evaluation of
        logical consistency.
  \item \textbf{FOLIO}~\citet{han2022folio} contains 1,430
        human-curated, open-domain statements with paired first-order-logic
        proofs. The task is to decide whether a conclusion is
        \textbf{Proved}, \textbf{Disproved}, or \textbf{Unknown}, which aligns well with FLD formats, making it a realistic testbed for natural-language reasoning.
  \item \textbf{ScienceQA}~\citet{lu2022learn} provides
        21,208 \textbf{Single-choice Science Questions} drawn
        from K–12 curricula. Roughly 49\% include an image context,
        48\% a text context, and 31\% both; every question has a
        grounded lecture and an explanation, facilitating evaluation of single-choice reasoning and interpretability.%
  \item \textbf{MMLU}~\citet{hendryckstest2021} is the standard
        \textbf{Massive Multitask Language Understanding} benchmark:
        15,908 four-choice multiple-choice questions spanning
        57 academic and professional disciplines. It measures a
        model’s broad knowledge and problem-solving ability in both zero-
        and few-shot settings.
  \item \textbf{MMLU-Pro}~\citet{wang2024mmluprorobustchallengingmultitask} extends the original
        MMLU benchmark by (i) filtering out trivial items and
        (ii) expanding each question’s answer set from 4 to
        10 options. The resulting collection 15,000
        questions across 57 subjects) substantially lowers random-guess
        accuracy and stresses multi-step reasoning.
    \item \textbf{CommonsenseQA}~\citet{talmor2019commonsenseqa} is a \textbf{Commonsense Multiple-Choice QA} benchmark built from ConceptNet relations. It contains 12,102 questions, each paired with 5 answer options (1 correct + 4 distractors) that are designed to be semantically plausible, making it a strong testbed for commonsense reasoning beyond surface word matching.
    
    \item \textbf{TruthfulQA}~\citet{lin2021truthfulqa} is a benchmark for evaluating \textbf{Truthfulness in QA} under common misconceptions. It includes 817 questions spanning diverse domains (e.g., health, law, finance, politics), and measures whether a model avoids producing answers that imitate widespread false beliefs, typically using both free-form generation and multiple-choice evaluation protocols.
    
    \item \textbf{MultiRC}~\citet{khashabi2018looking} is a \textbf{Multi-Sentence Reading Comprehension} dataset with questions that may have \textbf{multiple correct answers}. Each instance consists of a passage, a question, and a set of candidate answers, where the task is to judge each candidate as \textbf{True} (correct) or \textbf{False} (incorrect). This format targets compositional reasoning over several sentences and supports fine-grained evaluation of multi-answer understanding.
\end{itemize}

\begin{table*}[t]
  \centering
  \setlength{\tabcolsep}{6pt}
  \renewcommand{\arraystretch}{1.25}
  \caption{Comparison of \textbf{Knowledge Boundary} benchmarks for LLMs and MLLMs. Symbols denote levels of support (\cmark: fully implemented, \pmark: partially implemented, \xmark: not implemented).}
  \vspace{-5pt}
  \resizebox{\textwidth}{!}{%
  \begin{tabular}{l !{\vrule width 0.2pt} c c c !{\vrule width 0.2pt} c c c c !{\vrule width 0.2pt} c c c}
    \Xhline{1.5pt}
    \multirow{2}{*}{\raisebox{-1ex}{\textbf{\large Name}}}
      & \multicolumn{3}{c!{\vrule width 0.2pt}}{\rule{0pt}{3.2ex}\textbf{\large Coverage}}
      & \multicolumn{4}{c!{\vrule width 0.2pt}}{\rule{0pt}{3.2ex}\textbf{\large Performance}}
      & \multicolumn{3}{c}{\rule{0pt}{3.2ex}\textbf{\large Granularity}} \\
      & \makecell{Various\\Forms} & \makecell{Difficulty-\\Hierarchy} & \makecell{Supplement\\\& Expansion}
        & \makecell{Stability} & \makecell{General-\\ization} & \makecell{Bias\\Sensitivity} & Robustness
        & \makecell{External-\\Eval} & Holism & Diversity \\ [0.3ex]
    \midrule
\addlinespace
\textbf{\large TruthfulQA~\cite{lin2021truthfulqa}}
  & \xmark & \xmark & \pmark  & \pmark & \cmark & \pmark & \xmark & \xmark & \cmark & \pmark \\
\addlinespace
\textbf{\large POPE~\cite{li2023pope}}
  & \cmark & \xmark & \cmark  & \cmark & \cmark & \pmark & \pmark & \xmark & \xmark & \pmark \\
\addlinespace
\textbf{\large HaluEval~\cite{li2023haluevallargescalehallucinationevaluation}}
  & \xmark & \pmark & \cmark  & \xmark & \cmark & \pmark & \pmark & \xmark & \xmark & \pmark \\
\addlinespace
\textbf{\large Abstain-QA~\cite{madhusudhan2024abstainqa}}
  & \xmark & \pmark & \pmark  & \xmark & \cmark & \cmark & \xmark & \pmark & \xmark & \pmark \\
\addlinespace
\textbf{\large AMBER~\cite{wang2024amber}}
  & \cmark & \pmark & \cmark  & \cmark & \cmark & \cmark & \pmark & \cmark & \pmark & \cmark \\
\addlinespace
\textbf{\large HallusionBench~\cite{guan2024hallusionbench}}
  & \cmark & \pmark & \pmark  & \pmark & \cmark & \cmark & \pmark & \cmark & \pmark & \cmark \\
\addlinespace
\textbf{\large HaloQuest~\cite{wang2024haloquest}}
  & \cmark & \pmark & \cmark  & \pmark & \cmark & \pmark & \pmark & \pmark & \pmark & \cmark \\
\addlinespace
\textbf{\large LongHalQA~\cite{qiu2024longhalqa}}
  & \cmark & \pmark & \cmark  & \cmark & \cmark & \cmark & \pmark & \pmark & \xmark & \pmark \\
\addlinespace
\textbf{\large AttributionBench~\cite{li2024attributionbenchhardautomaticattribution}}
  & \xmark & \xmark & \xmark  & \cmark & \cmark & \pmark & \xmark & \cmark & \xmark & \xmark \\
\addlinespace
\textbf{\large RAGTruth~\cite{niu2024ragtruth}}
  & \xmark & \pmark & \pmark  & \xmark & \cmark & \pmark & \xmark & \pmark & \xmark & \pmark \\
\addlinespace
\textbf{\large HalluLens~\cite{bang2025hallulensllmhallucinationbenchmark}}
  & \xmark & \pmark & \cmark  & \cmark & \cmark & \cmark & \cmark & \pmark & \cmark & \cmark \\
\addlinespace
\textbf{\large VidHalluc~\cite{li2025vidhalluc}}
  & \cmark & \pmark & \cmark  & \cmark & \cmark & \cmark & \cmark & \pmark & \pmark & \cmark \\
\addlinespace
\textbf{\large PhD~\cite{liu2025phdchatgptpromptedvisualhallucination}}
  & \cmark & \pmark & \cmark  & \pmark & \cmark & \cmark & \pmark & \pmark & \cmark & \cmark \\
\addlinespace
\cdashline{1-11}
\addlinespace
\textbf{\textsc{\large BiasDetector} (Ours)}
  & \cmark & \cmark & \cmark  & \cmark & \cmark & \cmark & \cmark & \cmark & \cmark & \cmark \\
\addlinespace
\Xhline{1.5pt}
  \end{tabular}}
  \label{tab:new_benchmarks}
\vspace{-10pt}
\end{table*}

\begin{table*}[t]  
\centering
\caption{Percentage of \emph{Sparse Labels} (“Unknown”, “None Correct Option”) and the count of “Multiple Correct Options” across representative reasoning datasets. CommonsenseQA and TruthfulQA contain no \emph{Sparse Labels}.}
\label{Sparse Tense percentage}
\vspace{-5pt}

\renewcommand{\arraystretch}{1.3}
\setlength{\tabcolsep}{6pt}

\resizebox{\textwidth}{!}{%
\begin{tabular}{l l l r r r r}
\toprule
\textbf{Dataset} & \textbf{Question Type} & \textbf{Domain / Task} & \textbf{Total} & \textbf{Unknown (\%)} & \begin{tabular}[c]{@{}l@{}}\textbf{None Correct}\\\textbf{Option (\%)}\end{tabular} & \begin{tabular}[c]{@{}l@{}}\textbf{Multiple Correct}\\\textbf{Options}\end{tabular} \\
\midrule
\textit{FLD}~\cite{morishita2024enhancing} &
  \multirow{2}{*}{True / False / \textbf{Unknown}} &
  Logical (Deduction) &
  1\,100 & 100 (9.1) & --- & --- \\

\textit{FOLIO}~\cite{han2022folio} &  &
  Logical (Reasoning) &
  1\,430 & $\sim$470 ($\sim$33) & --- & --- \\
  
\textit{MMLU}~\cite{hendryckstest2021} &
  \multirow{4}{*}{Single-choice} &
  Multitask exam (57 subj.) &
  14\,042 & --- & 352 (2.5) & --- \\

\textit{ScienceQA}~\cite{lu2022scienceqa} & &
  Science QA (Single-choice) &
  21\,208 & --- & $\sim$500 ($\sim$2–3) & --- \\

\textit{CommonsenseQA}~\cite{talmor2019commonsenseqa} & &
  Commonsense reasoning &
  12\,102 & --- & 0 (0) & --- \\

\textit{TruthfulQA}~\cite{lin2021truthfulqa} & &
  Mis-/disinformation detection &
  817 & --- & 0 (0) & --- \\

\textit{MultiRC}~\citet{khashabi2018looking} &
  Multiple-choice (multi-ans.) &
  Multi-sentence reading comp. &
  9\,872 & --- & --- & 9\,872 \\  
\bottomrule
\end{tabular}%
}
\vspace{-10pt}
\end{table*}

\subsection{Labels Categorization Details}
As we introduced the ideation of \emph{Dense/Sparse Label} categorizations in Section~\ref{Label Categorization}, in this section we systematically analyze and survey the approximate percentage of samples contain \emph{Sparse Labels} take up in the whole field. As shown in Table~\ref{Sparse Tense percentage}, even though \emph{Sparse Labels} are not commonly seen in normal datasets, they still plays an important role in reasoning and thinking paradigms. Meanwhile, researching the generalization ability can be further used for user preference studying, as offering biased and flawed advice maybe included in users' prompt in wide applications.

\subsection{Sample Processing Details}
To construct the 4-choice, 2-answer multiple-choice questions from MultiRC~\citep{khashabi2018looking}, we extract questions and their answer options from the original dataset, where each answer is labeled with a binary \textbf{isAnswer} flag indicating correctness. We filter the dataset to retain only questions that have exactly four answer options and exactly two correct answers (i.e., two options with \textbf{isAnswer}=true). The options are then prefixed with numerical indices (0, 1, 2, 3) to standardize the format. This filtering process ensures that each question presents a balanced multiple-choice scenario with two correct options among four alternatives, suitable for evaluating models' ability to identify multiple correct answers simultaneously.

For \textbf{Judging datasets} like FLD and FOLIO, we directly extract 50\% of samples with \emph{True} and \emph{False} labels combined, and Unknown label for the rest 50\%. As the labeling are done by dataset creators, we did not introduce any bias. However, for \textbf{Single, None and Multi correct options MCQs}, we did no large revisions on samples themselves, only conservative labeling and option deleting, and applied manual verifications to ensure semantic coherence. Since we simulate the scenarios of low-quality datasets, we are creating samples that themselves containing biases that maybe hard to discover for users, so this setting is valid. We demonstrate samples in our used datasets in Appendix~\ref{Post-Processed Dataset Examples}.

\section{Related Work}
\label{related work}

As shown in table \ref{tab:new_benchmarks}, our proposed framework has comprehensively evaluates this phenomenon exceeding related work among \emph{eight} dimensions within three main categories, serving as an effective and comprehensive baseline for evaluating \emph{Instruction Boundary} phenomenon. From the experimental results, we derive empirical findings for these settings. Besides, we also carry out Benchmark-wise leader-board comparisons to demonstrate whether our findings are aligned with insights from other papers. Refer to Appendix \ref{related work} for detailed information.

\subsection{Introduction}
\paragraph{LLM Reasoning}
State-of-the-art reasoning stems from prompt-engineering advances such as \emph{chain-of-thought}~\cite{wei2022chain}, \emph{self-consistency}~\cite{wang2023selfconsistent}, program-aided logic (\textsc{PAL})~\cite{gao2023pal} and tool-augmented approaches like \textsc{ReAct}~\cite{yao2023react}.  
Datasets range from grade-school math GSM8K~\cite{cobbe2021gsm8k}, MATH~\cite{hendrycks2021math}, and \textsc{BBH}~\cite{suzgun2022bbh} to first-order-logic suites FLD/FOLIO and single-choice ScienceQA~\cite{lu2022scienceqa}.  
Our benchmark targets a complementary axis—\emph{instruction boundaries}—showing that even strong reasoners falter when critical options are omitted or contradictory.

\paragraph{LLM Trustworthiness}
Alignment pipelines span supervised fine-tuning, RLHF~\cite{ouyang2022instruct}, RLAIF in Llama-2~\cite{touvron2023llama2}, and reward-free DPO~\cite{rafailov2023dpo}.  
Safety-focused variants include many frameworks: iterative red-teaming~\cite{ganguli2022red}, \emph{Constitutional AI}~\cite{bai2022constitutionalaiharmlessnessai}, and inference-time filters such as \textsc{Llama-Guard}~\cite{inan2023llamaguardllmbasedinputoutput} or \textsc{SafeCompletion}~\cite{sun2023safecompletion}.  
Holistic evaluations (HELM~\cite{liang2022helm}, TrustLLM~\cite{cheema2024trustllm}) reveal residual hallucinations and over-confidence on benign inputs.  
We expose an orthogonal hazard: biased or erroneous answers triggered by \emph{incomplete instructions}, underscoring the need to couple alignment with coverage-aware calibration.

\paragraph{Hallucination \& Abstention}  
A growing body of benchmarks has emerged to evaluate hallucination detection and abstention quality in language models. Early efforts in this field such as \textsc{TruthfulQA}~\cite{lin2021truthfulqa} laid the foundation by probing model truthfulness, while subsequent benchmarks have expanded in scope and modality. For example, \textsc{POPE}~\cite{li2023pope} and \textsc{HaluEval}~\cite{li2023haluevallargescalehallucinationevaluation} explore model consistency and hallucination severity, while more recent datasets like \textsc{Abstain-QA}~\cite{madhusudhan2024abstainqa} and \textsc{AMBER}~\cite{wang2024amber} emphasize calibrated abstention and grounded reasoning. Other works such as the recent one \textsc{HallusionBench}~\cite{guan2024hallusionbench}, \textsc{HaloQuest}~\cite{wang2024haloquest}, and \textsc{LongHalQA}~\cite{qiu2024longhalqa} tackle domain-specific or long-context hallucination challenges. More recent contributions like \textsc{AttributionBench}~\cite{li2024attributionbenchhardautomaticattribution}, \textsc{RAGTruth}~\cite{niu2024ragtruth}, and \textsc{PhD} QA~\cite{liu2025phdchatgptpromptedvisualhallucination} continue to push the boundary of attribution and factual consistency, even extending to multimodal scenarios as seen in \textsc{VidHalluc}~\cite{li2025vidhalluc} and model introspection in \textsc{HalluLens}~\cite{bang2025hallulensllmhallucinationbenchmark}. Beyond supervised benchmarks, zero-resource methods such as \textsc{SelfCheckGPT}~\cite{manakul2023selfcheckgpt} offer scalable alternatives for automatic hallucination auditing without reliance on labeled samples.

\paragraph{Knowledge Boundary \& Uncertainty Analysis}
Early efforts framed the \emph{knowledge boundary} of LLMs as the parametric frontier beyond which hallucinations spike; recent work formalizes the concept and proposes dedicated probes.  \cite{li2024kbSurvey} synthesize taxonomy and open challenges, whereas \cite{yin2024benchmarking} and \cite{wen2024perception} construct benchmarks that search for prompt-sensitive knowledge limits.  Complementary methods such as projected-gradient attacks or semi-open questions further stress-test factual recall~\cite{zhang2024exploring}.  
On the orthogonal axis of \emph{predictive uncertainty}, empirical studies quantify epistemic and aleatoric risks in in-context learning~\cite{ling2024uncertaintyquantificationincontextlearning} and provide broader taxonomies of UQ techniques for LLMs~\cite{shorinwa2024uqSurvey}.  A recent position paper argues that classic dichotomies of uncertainty underestimate interactive agent settings~\cite{tranchero2025position}.  Our benchmark extends this line by unifying knowledge-coverage gaps with instruction incompleteness, exposing failure regimes where abstention should be favored over confident guesses.

\paragraph{Biases \& Conformity}
Beyond demographic stereotypes, LLMs inherit a spectrum of \emph{cognitive} and \emph{social} biases.  
Anchoring bias has been quantified across LLMs from different families like GPT-4, Claude-2, and Gemini-Pro~\cite{lou2024anchoringbiaslargelanguage}. Large-scale audits such as SALT~\cite{arif2024salt} and the Nature study on social-identity bias~\cite{hu2025social} show persistent preference gaps across gender, race, and religion. Under group pressure, single-turn experiments reveal majority‐aligned responses~\cite{zhu2024conformity}, while the multi-agent benchmark BenchForm~\cite{weng2025benchform} highlights escalation of conformity in these settings. These converging findings motivate our design: we examine how biased instructions magnifies anchoring and conformity, and evaluate calibration strategies that tackle knowledge gaps \emph{and} bias simultaneously.

\paragraph{LLM Question Answering}
Recent work has begun to move beyond single-answer MCQ evaluation by explicitly targeting
\emph{select-all-that-apply} (SATA) settings, where multiple options may be correct.
For example, SATA-BENCH shows that even strong LLMs struggle with multi-answer selection due to
systematic choice and count biases, and proposes decoding-time mitigation for more complete selection
\citep{xu2025satabenchselectapplybenchmark}.
In parallel, several studies reveal that LLM performance on MCQs is highly sensitive to
\emph{option-level perturbations}, such as reordering choices \citep{pezeshkpour2023largelanguagemodelssensitivity}
and even the evaluation protocol itself \citep{molfese2025rightanswerwrongscore}.
More directly aligned with our focus on imperfect option sets, recent work studies scenarios where
\emph{all provided options are incorrect}, highlighting the tension between helpfulness and critical rejection
(termed reflective judgment) \citep{góral2025waitthatsoptionllms}.
While these lines acknowledge that option completeness and correctness can substantially affect model behavior,
they typically examine a single failure mode in isolation.
In contrast, our work provides a controlled, unified framework that systematically varies
\emph{label sparsity} and \emph{option completeness} (complete vs.\ insufficient/redundant),
and evaluates models across heterogeneous question formats (e.g., per-candidate verification vs.\ MCQs),
thereby enabling direct comparisons of robustness and bias under realistic instruction defects.

\paragraph{Comparisons with Related Work}
The proposed “Instruction Boundary” taxonomy classifies prompts into sufficient, redundant, and insufficient categories. However, the paper does not compare this framework with existing research on label bias and calibration. For example, ~\cite{fei2023mitigatinglabelbiasesincontext} systematically categorized biases in in-context learning (ICL) into \emph{vanilla-label}, \emph{context-label}, and \emph{domain-label}, and proposed a \emph{domain-context calibration} method using randomly sampled domain words to quantify such biases. Their experiments showed that label bias can degrade LLM performance to near-random levels, while calibration significantly improves \cite{yu2024largelanguagemodelslearn,xu2025selfensemblemitigatingconfidencemiscalibration} further explain majority label bias and recency bias in ICL from the perspective of attention mechanisms. They found that similarity computations in the query-key matrix lead to a preference for frequent labels and proposed two simple methods—attention reweighting and positional embedding removal—that reduce majority and recency bias by 22\% and 17\%, respectively. In addition, the paper overlooks recent progress in confidence calibration. Xu et al. (2025) observed that LLMs suffer from severe confidence mis-calibration in multiple-choice QA—being under-confident in correct answers and over-confident in wrong ones—and proposed a Self-Ensemble method that groups and ensembles option predictions to mitigate this issue without extra data. ~\cite{zhang2024calibratingconfidencelargelanguage} decomposed confidence into “question uncertainty” and “answer faithfulness,” proposing the \emph{UF Calibration} framework that significantly improves calibration in RLHF models across four datasets:contentReference[oaicite:3]{index=3}. 

Regarding multi-answer evaluation, SATA-Bench~\citep{xu2025satabenchselectapplybenchmark} shows that existing LLMs achieve only about 41.8\% accuracy on “Select All That Apply” (SATA) tasks, with performance gaps mainly due to \emph{selection bias} and \emph{count bias}. Their “Choice Funnel” decoding strategy boosts accuracy by around 29\%. Lastly, ~\cite{baumann2025largelanguagemodelhacking} quantified “LLM hacking” risks in annotation pipelines, showing that small variations in model selection or prompt phrasing can induce 31–50\% false positive or false negative annotations, calling for strict verification when using LLMs as labelers. These works directly relate to prompt-induced biases and labeling risk; failure to discuss or benchmark against them weakens the novelty and clarity of the present contribution.

\section{Experiment Details}
\label{Biases_Details}
In this part, we introduce how the prototype of each setting come about, and how we systematically warp them up into the framework. Detailed prompt templates are available in Appendix~\ref{Settings Prompt}.

\noindent$\triangleright$ \textbf{Redundant Instructions:}
The phenomenon of \emph{addition bias} shows that LLMs often generate more content than necessary, leading to verbose or rambling outputs. Santagata and De~Nobili systematically analyzed this bias and observed that models trained on large text corpora tend to produce answers much longer than required across a variety of tasks, even when shorter answers suffice~\cite{santagata2024additionbias}. Such redundant instructions can interfere with reasoning by introducing irrelevant details, highlighting the need for techniques that encourage concise generation.

\noindent$\triangleright$ \textbf{Missing Option:}
Data contamination and memorization can cause LLMs to infer missing options even when the correct answer is absent. A protocol called \emph{Testset Slot Guessing} masks one option in a multiple-choice question and asks the model to fill the gap; commercial models correctly guessed the missing guidance more than half the time~\cite{anonymous2023datacontamination}. Separate work shows that when all presented options are incorrect, LLMs often still pick an answer and performance drops dramatically~\cite{goral2024wrongoptions}. These findings suggest that models may rely on spurious patterns rather than recognizing when no option applies.

\noindent$\triangleright$ \textbf{Vagueness:}
LLMs are acutely sensitive to prompt phrasing. The RobustAlpacaEval benchmark reveals that semantically equivalent prompts can yield wide performance swings; for instance, Llama‑2‑70B‑chat’s worst prompt accuracy was over 45 points lower than its best~\cite{cao2024worstprompt}. Minor alterations in format without semantic change can substantially degrade performance, and existing prompt-engineering techniques offer limited relief in these worst cases. Complementary work shows that only half of the performance gains from upgrading models stem from the new models themselves; the other half comes from users refining their prompts~\cite{murray2025promptstudy}. Moreover, automatically rewriting prompts can backfire by inserting unintended details and reducing task quality~\cite{murray2025promptstudy}. Together, these studies emphasize that vague or poorly specified instructions can distort reasoning accuracy and that prompt clarity is crucial.

\noindent$\triangleright$ \textbf{Few-shot Learning:}
In-context learning with few examples can induce predictive bias. Ma et~al.\ show that few-shot prompts exhibit high instability: performance varies with the choice, order and formatting of examples~\cite{ma2023fairnessfewshot}. They introduce a metric to quantify the predictive bias of a prompt and demonstrate that prompts with higher bias lead to worse predictive quality~\cite{ma2023fairnessfewshot}. A greedy search over prompt candidates can identify lower-bias prompts and improve few-shot performance. These results highlight that LLMs may favor labels appearing in the few-shot examples and underscore the need for balanced, representative demonstrations.

\noindent$\triangleright$ \textbf{Conformity:}
LLMs replicate human-like response biases such as recency and thus favor the last option presented—a recency bias—even when they vary paraphrasing or ordering~\cite{rupprecht2025promptperturb}. This bias persists across model sizes and underscores the importance of prompt design and robustness testing~\cite{rupprecht2025promptperturb}. Conformity also arises when models align with majority or initial cues; addressing conformity requires benchmarks and mitigation strategies that explicitly probe positional biases and anchoring effects~\cite{lou2024anchoringbiaslargelanguage,zhu2024conformity,shoval2025conformityTrial,weng2025benchform}.

\noindent$\triangleright$ \textbf{Prompt Polishing:}
Hidden or rewritten prompts can introduce unintended biases. Neumann et~al.\ highlight that system-level directives—unseen by end users—take precedence over user instructions and can create representational and assign biases across demographic groups~\cite{neumann2025systemprompts}. Because deployments can append additional system prompts without mutual visibility, these layers of instructions may interact in unpredictable ways and produce downstream harms~\cite{neumann2025systemprompts}. Similarly, automatic prompt rewriting intended to help users can conflict with their intent and degrade performance by adding extraneous details~\cite{murray2025promptstudy}. These results caution against naive prompt polishing and motivate transparent, user-controlled instruction design.

\noindent$\triangleright$ \textbf{Reasoning-Reflection:}
Reminding the model of previous wrong answering to reduce bias through LLM reflection. The B-score metric compares the distribution of answers in single-turn and multi-turn conversations; biases such as a preference for a particular number diminish when the model takes its own history into account~\cite{vo2025bscore}. For instance, GPT‑4o strongly favors the number 7 in single-turn random-choice questions, but in multi-turn settings it alternates answers and achieves a uniform distribution~\cite{vo2025bscore}. This unsupervised approach requires no labeled data and demonstrates that maintaining response history enables models to self-correct and mitigate biases. Such findings suggest that multi-turn dialogue and explicit reflection can be leveraged to improve calibration and fairness in LLMs.

\begin{table*}[t]
\centering
\renewcommand{\arraystretch}{1.2}
\caption{Detailed Accuracy of the left picture in Figure~\ref{fig:side results}, 
showing LLMs’ robustness against \textbf{Disturbing Miscellany} setting. 
Highest values are marked darker.}
\label{tab:misc-acc}
\vspace{-5pt}
\resizebox{\textwidth}{!}{%
\begin{tabular}{l c c c c c c c c c}
\toprule
\multirow{2}{*}{\textbf{Model}} & \multicolumn{9}{c}{\textbf{Accuracy regarding False Choices Numbers (\%)}} \\
\cmidrule(lr){2-10}
 & \textbf{1} & \textbf{2} & \textbf{3} & \textbf{4} & \textbf{5} & \textbf{6} & \textbf{7} & \textbf{8} & \textbf{9} \\
\midrule
gpt-3.5-turbo         & 84.00 & 84.33 & 77.33 & 75.67 & 74.33 & 76.33 & 73.67 & 73.00 & 71.33 \\
gpt-4o                & 98.67 & 97.00 & 96.33 & 97.00 & 97.00 & 97.33 & 96.00 & 96.00 & 96.67 \\
llama-3.1-8b & 96.67 & 95.33 & 96.33 & 95.00 & 95.67 & 95.33 & 95.00 & 94.00 & 95.33 \\
claude-3.7-sonnet     & \textbf{100.00} & \textbf{99.67} & \textbf{99.67} & \textbf{100.00} & 98.33 & \textbf{99.00} & \textbf{99.33} & 98.67 & \textbf{98.33} \\
gemini-2.0-flash      & \textbf{100.00} & 98.67 & 98.67 & 98.33 & \textbf{99.33} & 98.33 & 98.33 & \textbf{98.67} & 97.33 \\
\bottomrule
\end{tabular}
}
\end{table*}

\section{Experiment Details}
\subsection{Output Identifications}
\paragraph{Keyword Matching List}
As shown in Appendix~\ref{keyword_lists}, we first employ keyword matching to categorize the outputs. Most of them are related synonyms differences, and this process covers more than 95
For the remaining exceptional cases, we apply LLM-as-a-Judge to determine the label, similar to approaches used in~\cite{zhou2025dissectinglogicalreasoningllms}.
Moreover, recent work on multilingual robustness such as \citet{li-etal-2025-multilingual-federated} further highlights that annotation frameworks must remain reliable under multilingual or cross-lingual variations, suggesting that judge-based evaluation can also benefit from disentangled linguistic representations.
Detailed results can be found in Table~\ref{tab:judge-usage}.

\paragraph{LLM-as-a-Judge}
\label{LLM Judge Verifications}
In our experiment, we apply Qwen-2.5-7b as our general LLM Judge. As indicated in~\cite{chehbouni2025validreliableinvestigatinguse}, LLM-as-a-Judge can be potentially invalid and unreliable, we apply \textbf{Testing Cases Analysis}. Qwen has passed \textbf{ALL} the testing cases to prove its precision and reliability, detailed can be  in Appendix~\ref{LLM Judge Testing Cases}. Meanwhile, we count the percentage of the Judge triggered as the keyword matching failed. Most of the tasks showed less than 5\% LLM-as-a-Judge usage, the risk is relatively low; Besides, we apply Grok3-mini as another Judge model, both results show unified judgments and passed all the testing cases, proving our evaluation method is valid. LLM Judge templates are in Appendix~\ref{LLM Judge Templates}.

\begin{table*}[t]
\centering
\caption{LLM-As-A-Judge usage ratio across different experiment settings, 
which also denotes the percentage of LLM output contains neither of the words in Appendix~\ref{keyword_lists}.}
\label{tab:judge-usage}
\vspace{-5pt}
\resizebox{\textwidth}{!}{  
\begin{tabular}{@{}c c c c c c c c@{}}  
\toprule
\textbf{Models} & \makecell{\textbf{Vanilla} \\ \textbf{Scenario}} 
& \makecell{\textbf{Missing-} \\ \textbf{Choices}} 
& \makecell{\textbf{Few-shot} \\ \textbf{Learning}} 
& \makecell{\textbf{Vagueness}} 
& \makecell{\textbf{Prompt-} \\ \textbf{Polishing}} 
& \makecell{\textbf{Multi-turn} \\ \textbf{Dialogue}} 
& \makecell{\textbf{Conformity}} \\
\midrule
\textbf{GPT-3.5-turbo} & 26.44 & 18.89 & 16.22 & 10.52 & 22.67 & 26.44 & 1.78 \\
\textbf{GPT-4o} & 4.44 & 6.22 & 4.44 & 6.67 & 3.56 & 6.44 & 3.33 \\
\textbf{Llama-3.1-8b} & 10.44 & 12.44 & 10.89 & 10.67 & 35.78 & 35.33 & 7.11 \\
\textbf{Claude-3.7-Sonnet} & 35.11 & 34.67 & 24.22 & 28.89 & 35.56 & 16.44 & 36.44 \\
\textbf{Gemini-2.0-flash} & 4.00 & 8.00 & 7.11 & 10.67 & 3.11 & 9.33 & 6.00 \\
\bottomrule
\end{tabular}
}
\end{table*}

\section{MCQs Answer Validation Logic Verification}

To ensure the accuracy of multiple-choice answer validation logic in our Choice experiments, we have designed a comprehensive test suite to verify the correctness of our evaluation algorithm, as shown in Appendix~\ref{Multi-Choices Testing Cases}. This section provides detailed documentation of the validation principles, test cases, and results.

\subsection{Core Validation Principles}

The multiple-choice validation logic is based on the following fundamental principles:

\begin{enumerate}
    \item \textbf{Strict Matching Principle}: Multiple-choice answers must exactly match the ground truth, requiring all correct options to be selected without any incorrect options.
    \item \textbf{Order Independence}: Option order does not affect the validation result, ensured through sorting operations.
    \item \textbf{Complete Coverage}: Partial correctness (missing correct options or including wrong options) is treated as incorrect.
\end{enumerate}

The core validation principle is that multiple-choice answers must exactly match the ground truth through strict comparison of sorted option lists.

\subsection{Test Case Design}

We have designed comprehensive test cases covering various scenarios:

\begin{itemize}
    \item \textbf{Perfect Match Tests}: Verify recognition of correct answers in different formats
    \item \textbf{Partial Correctness Tests}: Verify rejection of partially correct answers
    \item \textbf{Extra Options Tests}: Verify rejection of answers containing additional options
    \item \textbf{Completely Wrong Tests}: Verify rejection of completely incorrect answers
    \item \textbf{No Answer Tests}: Verify handling of no-answer scenarios
    \item \textbf{Edge Case Tests}: Verify handling of exceptional inputs
\end{itemize}

\begin{table}[t]                         
\centering
\caption{Validation Accuracy by Test Category}
\label{tab:validation_accuracy}

\tiny
\resizebox{\columnwidth}{!}{%
\begin{tabular}{@{}lcc@{}}
\toprule
\textbf{Test Category} & \textbf{Passed} & \textbf{Accuracy} \\
\midrule
Perfect Match        & 6/6  & 100\% \\
Partial Correctness  & 4/4  & 100\% \\
Extra Options        & 3/3  & 100\% \\
Completely Wrong     & 4/4  & 100\% \\
No Answer            & 5/5  & 100\% \\
Edge Cases           & 5/5  & 100\% \\
\midrule
\textbf{Overall}     & \textbf{27/27} & \textbf{100\%} \\
\bottomrule
\end{tabular}%
}
\end{table}

\begin{table*}[!t]
  \centering
  \caption{Experiment Results in our proposed prompt templates, and Accuracy comparisons with the \emph{Vanilla} setting.}
  \label{main results}
  \vspace{-5pt}
  \small
  \resizebox{\textwidth}{!}{%
  \begin{tabular}{@{}l c c c : c c c c@{}}
    \toprule
    & \multicolumn{3}{c}{\textbf{Judging}} &
      \multicolumn{4}{c}{\textbf{MCQs}} \\
    \cmidrule(lr){2-4}\cmidrule(lr){5-8}
    \textbf{Models} & \textbf{$Acc$} & \textbf{$SR$} & \textbf{$GR$} &
            \textbf{$Acc$} & \textbf{$SR$} & \textbf{$GR^1$} & \textbf{$GR^2$} \\
    \midrule

\rowcolor{gray!20}
    \makecell{GPT-3.5-turbo}
      & \begin{tabular}[t]{@{}c@{}}57.50\\[-0.6ex]\textcolor{darkgreen}{\scriptsize($\uparrow$ 8.75)}\end{tabular}
      & 35.00 & 80.00
      & \multicolumn{1}{|c}{\begin{tabular}[t]{@{}c@{}}54.44\\[-0.6ex]\textcolor{darkgreen}{\scriptsize($\uparrow$ 29.77)}\end{tabular}}
      & 35.33 & 81.33 & 46.67 \\

    \makecell{GPT-4o}
      & \begin{tabular}[t]{@{}c@{}}38.67\\[-0.6ex]\textcolor{red}{\scriptsize($\downarrow$ 6.74)}\end{tabular}
      & 54.67 & 22.67
      & \multicolumn{1}{|c}{\begin{tabular}[t]{@{}c@{}}64.00\\[-0.6ex]\textcolor{darkgreen}{\scriptsize($\uparrow$ 24.67)}\end{tabular}}
      & 68.00 & 85.33 & 38.67\\

\rowcolor{gray!20}
    \makecell{LLaMA-3.1-8B}
      & \begin{tabular}[t]{@{}c@{}}33.83\\[-0.6ex]\textcolor{red}{\scriptsize($\downarrow$ 18.04)}\end{tabular}
      & 31.67 & 36.00
      & \multicolumn{1}{|c}{\begin{tabular}[t]{@{}c@{}}23.56\\[-0.6ex]\textcolor{darkgreen}{\scriptsize($\uparrow$ 2.89)}\end{tabular}}
      & 40.00 & 20.67 & 10.00 \\

    \makecell{Claude-3.7-Sonnet}
      & \begin{tabular}[t]{@{}c@{}}72.67\\[-0.6ex]\textcolor{darkgreen}{\scriptsize($\uparrow$ 26.44)}\end{tabular}
      & 66.33 & 79.00
      & \multicolumn{1}{|c}{\begin{tabular}[t]{@{}c@{}}59.11\\[-0.6ex]\textcolor{darkgreen}{\scriptsize($\uparrow$ 51.33)}\end{tabular}}
      & 64.67 & 56.00 & 56.67 \\

\rowcolor{gray!20}
    \makecell{Gemini-2.0-Flash}
      & \begin{tabular}[t]{@{}c@{}}67.17\\[-0.6ex]\textcolor{darkgreen}{\scriptsize($\uparrow$ 15.36)}\end{tabular}
      & 53.33 & 81.00
      & \multicolumn{1}{|c}{\begin{tabular}[t]{@{}c@{}}52.89\\[-0.6ex]\textcolor{darkgreen}{\scriptsize($\uparrow$ 17.89)}\end{tabular}}
      & 45.33 & 66.67 & 46.67 \\

    \bottomrule
  \end{tabular}%
}
\vspace{-5pt}
\end{table*}

\begin{table}[t]
\centering
\caption{Significance Testing comparing biased settings with the \emph{Vanilla} setting.
. Each model is treated as an independent observation, all tested models are considered.
$\Delta Acc = Acc_{b} - Acc_{\text{v}}$.}
\label{tab:significance_test}
\small
\setlength{\tabcolsep}{6pt}
\renewcommand{\arraystretch}{1.15}
\resizebox{1.0\linewidth}{!}{%
\begin{tabular}{l c c}
\toprule
\textbf{Setting} & \textbf{\textbf{$\bar{\Delta}$Acc}} & \textbf{$p$-value} \\
\midrule
Reasoning-reflection   & $-8.10$ & $\mathbf{0.002}$ \\
Prompt-Polishing  & $-5.94$ & $0.089$ \\
Conformity        & $+3.83$ & $0.20$ \\
Few-shot Learning & $+3.39$ & $0.19$ \\
Missing Choices   & $-4.01$ & $\mathbf{0.030}$ \\
Vagueness         & $-8.72$ & $\mathbf{0.012}$ \\
Disturbing Miscellany & $8.18$ & $\mathbf{0.00029}$ \\
\bottomrule
\end{tabular}%
}
\end{table}

\subsection{Test Results}

We present detailed test results using the template:

\subsection{Test Results Analysis}

The test results demonstrate that our multiple-choice validation logic performs correctly across all test cases:

\begin{itemize}
    \item \textbf{Perfect Match Recognition}: Successfully identifies various formats of completely correct answers, including comma-separated, and-connector, space-separated, and other formats.
    \item \textbf{Partial Correctness Rejection}: Correctly rejects partially correct answers, ensuring strict standards for multiple-choice questions.
    \item \textbf{Extra Options Rejection}: Correctly rejects answers containing additional options, preventing over-selection.
    \item \textbf{Completely Wrong Rejection}: Correctly rejects completely incorrect answers.
    \item \textbf{No Answer Handling}: Properly handles various no-answer scenarios.
    \item \textbf{Edge Case Robustness}: Successfully processes exceptional inputs without errors.
\end{itemize}

\subsection{Algorithm Complexity Analysis}

The validation logic has a time complexity of $O(n \log n)$, where $n$ is the number of options, primarily due to sorting operations. The space complexity is $O(n)$ for storing the duplicated option list. For 4-option multiple-choice questions, the algorithm efficiency fully meets real-time requirements.

\subsection{Supported Response Formats}

Our validation system supports a wide range of response formats:

\begin{itemize}
    \item \textbf{Comma-separated}: \texttt{1, 3}, \texttt{1,3}
    \item \textbf{And-connector}: \texttt{1 and 3}
    \item \textbf{Space-separated}: \texttt{1 3}
    \item \textbf{Single option}: \texttt{1}
    \item \textbf{No answer indicators}: \texttt{No correct answer}, \texttt{uncertain}, \texttt{unclear}
\end{itemize}

\subsection{Conclusion}

Through comprehensive testing and verification, we confirm that our multiple-choice validation logic accurately and strictly evaluates the correctness of multiple-choice answers, providing a reliable foundation for Choice experiments. This ensures fairness and accuracy in multiple-choice evaluation, meeting rigorous requirements of academic research.

The validation system demonstrates: (1) 100\% accuracy across all test categories. (2) Robust handling of various input formats. (3) Strict standards for answer evaluation. (4) Comprehensive coverage of edge cases.

This validation logic provides a solid foundation for our Choice experiments, ensuring that model performance is evaluated fairly and accurately across all question types.

\section{Problem Mitigation Details}
\label{Proposed Template}
We design prompt templates for both \emph{Judging question} and \emph{MCQs} covering all biases in Appendix~\ref{Problem Mitigation}. Experiments show that most models show a higher accuracy, although GPT-4o and LLaMA-3.1-8b had a accuracy decline, the distributions are more even, as shown in Table~\ref{main results}, which proves our templates could serve as a prototype of problem mitigation.

\section{Significance Testing}
\label{Significance Testing}

To examine whether instruction settings induce systematic effects beyond individual models, we conduct significance testing by treating each model as an independent observation and comparing its performance against the Vanilla baseline.
As shown in Table~\ref{tab:significance_test}, we calculate the average accuracy and the p-values of all biased settings compared with the \emph{Vanilla} setting. Across five models, we observe that several settings lead to consistent and statistically reliable changes in judging accuracy.
In particular, \emph{Reasoning-reflection}, \emph{Missing Choices}, and \emph{Vagueness} significantly degrade performance across models, indicating that these interventions introduce stable disruptions to the decision process.
In contrast, \emph{Conformity} and \emph{Few-shot Learning} exhibit positive average improvements; however, their effects are not statistically significant due to large variability across models.
These results suggest that instruction-level modifications do not uniformly improve accuracy and that apparent gains may be fragile and model-dependent.
Overall, the analysis highlights the risk of relying on prompt-based strategies without accounting for their cross-model stability. 

The used formula and calculation steps are as follows:

\begin{equation}
\bar{x} = {\overline \Delta Acc}
= \frac{1}{n}\sum_{i=1}^{n}\Delta Acc_i
\end{equation}
\noindent\textit{Here, $\bar{x}$ denotes the sample mean of the accuracy differences relative to the baseline setting.}

\begin{equation}
s = \sqrt{\frac{1}{n-1}\sum_{i=1}^{n}\left(\Delta Acc_i - \bar{x}\right)^2}
\end{equation}
\noindent\textit{This equation computes the sample standard deviation of the accuracy differences.}

\begin{equation}
SE = \frac{s}{\sqrt{n}}
\end{equation}
\noindent\textit{The standard error reflects the uncertainty of the estimated mean difference.}

\begin{equation}
t_{\text{obs}} = \frac{\bar{x} - \mu_0}{s/\sqrt{n}}
\end{equation}
\noindent\textit{The one-sample $t$ statistic compares the observed mean difference against the hypothesis mean $\mu_0$.}

\begin{equation}
t_{\text{obs}} = \frac{\bar{x}}{s/\sqrt{n}}
\end{equation}
\noindent\textit{In our setting, the null hypothesis assumes $\mu_0 = 0$.}

\begin{equation}
df = n - 1
\end{equation}
\noindent\textit{The degrees of freedom.}

\begin{equation}
T_{df} \sim t(df)
\end{equation}
\noindent\textit{Under the hypothesis, the test statistic follows a Student’s $t$ distribution with $df$ degrees of freedom.}

\begin{equation}
p = \Pr\!\left(\left|T_{df}\right| \ge \left|t_{\text{obs}}\right|\right)
\end{equation}
\noindent\textit{The two-sided $p$-value is defined as the probability of observing a statistic at least as extreme as $t_{\text{obs}}$.}

\section{Few-shot Learning Biases Details}
\label{Few-shot details}
In the main body, we only analysis few-shot learning containing only one examples. However, we also carry out comparisons regarding different label types and numbers, we observe that when each of the three labeled samples are included as Few-shot learning examples, model’s accuracy actually increases when we remove Unknown-labeled samples and only include \emph{True} and \emph{False} labeled samples each, but times of LLMs outputting \emph{Unknown} increases—yet the overall accuracy is slightly higher than all-three-label contained Few-shot condition. \textbf{Bringing uncertainty samples with \emph{Unknown} label is generally harmful or useless to overall reasoning performance. Uncertainty estimation ability cannot be improved via sole imitations.}

\section{LLM Usage Statement}
GPT-4o, o3 and 5-Thinking are used to help polish up paper writing and statistical charts. Claude-Sonnet-4.5 and Grok Code are used for Code framing and toolkit organizing.

\section{Knowledge Conflict Analysis}
\label{Knowledge Conflict}
Resistance against these redundant can be different, we consider this phenomenon attributed to \emph{Knowledge Conflict}: based on the model's original output distributions, conformity of \emph{Dense Labels} is more likely to demonstrates higher robustness, on the other hand, \emph{Sparse Labels} can trigger conflicts seriously. Here we apply same experiment settings of \emph{Few-shot Learning} and \emph{Conformity} regarding \emph{Dense Labels} to test the biases of LLMs. As depicted in Figure~\ref{fig:side results}, we observe that the accuracy varies, indicating LLMs exhibit a systematic bias principles regarding different imitation aspects. They cognitively handle different types of labels. This pattern holds true regardless of the missing label types. \textbf{Robustness of LLM Reasoning is not label-isotropic but label-sensitive. Confronting \emph{Sparse Labels}, models are more likely to show flaws.}

\section{Potential Risks Discussion}
Our experiments are conducted in strict accordance with the ethical guidelines, we acknowledge that our study does not include a thorough analysis of datasets that may raise ethical concerns. This limitation stems from our focus on publicly available and widely accepted benchmarks, which, despite their prevalent use, might still contain biases, privacy issues, or other subtle ethical implications that are not immediately apparent.
While our study focuses on evaluating LLMs using mixed question forms and sparsely-labeled datasets—many of which reflect realistic yet low-quality input scenarios—enhancing LLMs' ability to handle such inputs may inadvertently amplify risks of dataset misuse. For instance, if LLMs become proficient at confidently answering under-specified, inconsistent, or misleading questions, they may propagate unreliable information in high-stakes applications such as education, healthcare, or policy-making. Furthermore, models might inadvertently overfit to or memorize common artifacts from benchmark datasets if these formats resemble training data, leading to an overestimation of generalization.
\clearpage  
\begin{figure*}[t]
\centering
\caption{Detailed output distribution of LLMs on judging‐ and choice‐style datasets. Highest values per column are in \textbf{bold}.}
\label{tab:merged-output-distribution}
\vspace{-6pt}
\renewcommand{\arraystretch}{1.15}
\setlength{\tabcolsep}{6pt}
\small
\resizebox{\textwidth}{!}{%
\begin{tabular}{@{}l c c c c c c c@{}}
  \toprule
  \textbf{Model} &
  \textbf{Unknown (\%)} & \textbf{True (\%)} & \textbf{False (\%)} &
  \makecell[c]{\textbf{Single}\\\textbf{Option (\%)}} &
  \makecell[c]{\textbf{Multiple}\\\textbf{Options (\%)}} &
  \makecell[c]{\textbf{None}\\\textbf{Option (\%)}} \\
  \midrule
  GPT-3.5-turbo         & 47.81 & 48.90 &  3.30 & 49.1 & 19.6 & 31.3 \\
  GPT-4o                & 22.19 & \textbf{77.18} &  0.06 & \textbf{66.0} & 19.8 & 14.2 \\
  Llama-3.1-8b & \textbf{69.22} & 30.78 &  0.00 & 52.0 & 39.6 &  8.4 \\
  Claude-3.7-sonnet     & 18.75 & 72.03 &  9.22 & 24.7 & \textbf{41.8} & \textbf{33.6} \\
  Gemini-2.0-flash      & 52.34 & 36.41 & \textbf{11.25} & 61.0 & 28.0 & 11.0 \\
  \bottomrule
\end{tabular}}
\vspace{-6pt}
\end{figure*}

\vspace{10pt} 

\begin{figure*}[t]
\centering
\caption{Detailed values of Robustness Score (RS) for all main settings across five LLMs (higher is better). Highest values in \textbf{bold}.}
\label{tab:detailed-rs}
\setlength{\tabcolsep}{9pt}
\renewcommand{\arraystretch}{1.2}
\resizebox{\textwidth}{!}{%
\begin{tabular}{lccccccc}
\toprule
\textbf{Model} &
\makecell[c]{Vanilla\\Scenario} &
\makecell[c]{Missing\\Option} &
\makecell[c]{Few\\shot} &
\makecell[c]{Vagueness} &
\makecell[c]{Prompt\\Polishing} &
\makecell[c]{Multi\\turn} &
\makecell[c]{Con\\formity} \\
\midrule
gpt-3.5-turbo         & 33.48 & 14.89 &  1.24 & 31.74 & 15.87 & 35.55 & 10.10 \\
gpt-4o                & 38.20 & \textbf{42.35} & 45.31 & 20.88 & \textbf{46.08} & 37.30 & 25.63 \\
llama-3.1-8b & 29.65 & 15.76 & 25.34 &  0.63 & 41.18 & 29.63 &  0.00 \\
claude-3.7-sonnet     & 45.42 & 39.62 & \textbf{54.09} & \textbf{50.53} & 41.62 & \textbf{43.65} & \textbf{31.05} \\
gemini-2.0-flash      & \textbf{47.57} & 41.21 & 30.32 & 29.65 &  1.48 & 40.81 &  8.93 \\
\bottomrule
\end{tabular}%
}
\end{figure*}






\clearpage 
\onecolumn 

\section{LLM Judge Testing Cases}
\label{LLM Judge Testing Cases}

\raggedbottom

\setlist[itemize]{nosep,topsep=2pt,itemsep=1pt,parsep=0pt,leftmargin=1.4em}

\newcommand{\mytcolorbox}[2]{%
  \par\noindent
  
  \begin{tcolorbox}[
    colback=white,
    colframe=green!40!black,
    title=#1,
    coltitle=white,
    colbacktitle=green!40!black,
    fonttitle=\bfseries,
    width=\textwidth,
    boxrule=1pt,
    arc=4pt,
    boxsep=4pt,
    left=8pt,right=8pt,top=6pt,bottom=6pt,
    before skip=6pt,
    after skip=6pt
  ]
  #2
  \end{tcolorbox}
}


\mytcolorbox{Emotional Expressions}{
\textbf{Description:} Responses expressing emotional states or personal opinions

\textbf{Support Cases:}
\begin{itemize}
\item \textbf{Description:} Emotional support
\item \textbf{Response:} "I'm confident that the evidence strongly confirms this conclusion."
\item \textbf{Expected:} \_\_PROVED\_\_ \textbf{LLM Judge Result:} \_\_PROVED\_\_ 
\end{itemize}

\textbf{Uncertainty Cases:}
\begin{itemize}
\item \textbf{Description:} Emotional uncertainty
\item \textbf{Response:} "I'm hesitant about the implications of this evidence."
\item \textbf{Expected:} \_\_UNKNOWN\_\_ \textbf{LLM Judge Result:} \_\_UNKNOWN\_\_ 
\end{itemize}
}

\mytcolorbox{Technical Expressions}{
\textbf{Description:} Responses using technical terminology and specialized language

\textbf{Support Cases:}
\begin{itemize}
\item \textbf{Description:} Technical support
\item \textbf{Response:} "The quantitative analysis yields results that are consistent with the proposed hypothesis."
\item \textbf{Expected:} \_\_PROVED\_\_ \textbf{LLM Judge Result:} \_\_PROVED\_\_ 
\end{itemize}

\textbf{Uncertainty Cases:}
\begin{itemize}
\item \textbf{Description:} Technical uncertainty
\item \textbf{Response:} "The sample size limitations prevent us from making robust statistical inferences."
\item \textbf{Expected:} \_\_UNKNOWN\_\_ \textbf{LLM Judge Result:} \_\_UNKNOWN\_\_ 
\end{itemize}
}

\mytcolorbox{Implicit Expressions}{
\textbf{Description:} Responses expressing support or uncertainty through implicit language

\textbf{Support Cases:}
\begin{itemize}
\item \textbf{Description:} Implicit support expression
\item \textbf{Response:} "The evidence aligns with the statement, suggesting a favorable outcome."
\item \textbf{Expected:} \_\_PROVED\_\_ \textbf{LLM Judge Result:} \_\_PROVED\_\_ 
\end{itemize}

\textbf{Opposition Cases:}
\begin{itemize}
\item \textbf{Description:} Implicit opposition expression
\item \textbf{Response:} "The data reveals inconsistencies that question the statement's soundness."
\item \textbf{Expected:} \_\_DISPROVED\_\_ \textbf{LLM Judge Result:} \_\_DISPROVED\_\_ 
\end{itemize}

\textbf{Uncertainty Cases:}
\begin{itemize}
\item \textbf{Description:} Implicit uncertainty expression.
\item \textbf{Response:} "The situation presents multiple variables that complicate a straightforward assessment."
\item \textbf{Expected:} \_\_UNKNOWN\_\_ \textbf{LLM Judge Result:} \_\_UNKNOWN\_\_ 
\end{itemize}
}

\mytcolorbox{Metaphors and Analogies}{
\textbf{Description:} Responses using metaphorical language and analogies to express positions

\textbf{Support Cases:}
\begin{itemize}
\item \textbf{Description:} Metaphorical support
\item \textbf{Response:} "Like a puzzle with all pieces fitting perfectly, the evidence confirms the statement completely."
\item \textbf{Expected:} \_\_PROVED\_\_ \textbf{LLM Judge Result:} \_\_PROVED\_\_ 
\end{itemize}

\textbf{Opposition Cases:}
\begin{itemize}
\item \textbf{Description:} Metaphorical opposition
\item \textbf{Response:} "The evidence is like a house of cards - it appears solid but collapses under pressure."
\item \textbf{Expected:} \_\_DISPROVED\_\_ \textbf{LLM Judge Result:} \_\_DISPROVED\_\_ 
\end{itemize}

\textbf{Uncertainty Cases:}
\begin{itemize}
\item \textbf{Description:} Metaphorical uncertainty
\item \textbf{Response:} "The situation is like a foggy landscape where the path forward remains indistinct."
\item \textbf{Expected:} \_\_UNKNOWN\_\_ \textbf{LLM Judge Result:} \_\_UNKNOWN\_\_ 
\end{itemize}
}

\mytcolorbox{Conditional Responses}{
\textbf{Description:} Responses that present conditional or hypothetical reasoning

\textbf{Support Cases:}
\begin{itemize}
\item \textbf{Description:} Conditional support
\item \textbf{Response:} "Under the current circumstances and given the available data, the statement appears to be sound."
\item \textbf{Expected:} \_\_PROVED\_\_ \textbf{LLM Judge Result:} \_\_PROVED\_\_ 
\end{itemize}

\textbf{Uncertainty Cases:}
\begin{itemize}
\item \textbf{Description:} Conditional response
\item \textbf{Response:} "If we consider the primary evidence, the statement holds merit. However, alternative interpretations exist."
\item \textbf{Expected:} \_\_UNKNOWN\_\_ \textbf{LLM Judge Result:} \_\_UNKNOWN\_\_ 
\item \textbf{Description:} Conditional uncertainty
\item \textbf{Response:} "While the initial assessment seems favorable, the underlying assumptions may not hold under scrutiny."
\item \textbf{Expected:} \_\_UNKNOWN\_\_ \textbf{LLM Judge Result:} \_\_UNKNOWN\_\_
\end{itemize}
}

\mytcolorbox{Academic Expressions}{
\textbf{Description:} Responses using formal academic language and terminology

\textbf{Support Cases:}
\begin{itemize}
\item \textbf{Description:} Academic support expression
\item \textbf{Response:} "The empirical findings demonstrate a statistically significant correlation that confirms the hypothesis."
\item \textbf{Expected:} \_\_PROVED\_\_ \textbf{LLM Judge Result:} \_\_PROVED\_\_
\end{itemize}

\textbf{Opposition Cases:}
\begin{itemize}
\item \textbf{Description:} Academic opposition expression
\item \textbf{Response:} "The methodology reveals fundamental flaws that undermine the credibility of the conclusions."
\item \textbf{Expected:} \_\_DISPROVED\_\_ \textbf{LLM Judge Result:} \_\_DISPROVED\_\_ 
\end{itemize}

\textbf{Uncertainty Cases:}
\begin{itemize}
\item \textbf{Description:} Academic uncertainty expression
\item \textbf{Response:} "The research design presents limitations that prevent definitive conclusions from being drawn."
\item \textbf{Expected:} \_\_UNKNOWN\_\_ \textbf{LLM Judge Result:} \_\_UNKNOWN\_\_
\end{itemize}
}

\mytcolorbox{Complex Reasoning Responses}{
\textbf{Description:} Responses involving complex logical reasoning without explicit keywords

\textbf{Support Cases:}
\begin{itemize}
\item \textbf{Description:} Complex reasoning support
\item \textbf{Response:} "Based on the comprehensive analysis of the provided evidence, the logical conclusion points toward the soundness of this statement."
\item \textbf{Expected:} \_\_PROVED\_\_ \textbf{LLM Judge Result:} \_\_PROVED\_\_ 
\end{itemize}

\textbf{Opposition Cases:}
\begin{itemize}
\item \textbf{Description:} Complex reasoning opposition
\item \textbf{Response:} "The evidence presented leads to a conclusion that questions the initial statement."
\item \textbf{Expected:} \_\_DISPROVED\_\_
\textbf{LLM Judge Result:} \_\_DISPROVED\_\_ 
\end{itemize}

\textbf{Uncertainty Cases:}
\begin{itemize}
\item \textbf{Description:} Complex reasoning uncertainty
\item \textbf{Response:} "The complexity of this situation requires deeper analysis and presents multiple perspectives."
\item \textbf{Expected:} \_\_UNKNOWN\_\_
\textbf{LLM Judge Result:} \_\_UNKNOWN\_\_ 
\end{itemize}
}


\section{Prompt Templates (MCQs Dataset)}
\label{sec:choice_prompts}

\begin{tcolorbox}[
    colback=orange!5!white,
    colframe=orange!75!black,
    title=Vanilla Scenario (MCQs)
]
\texttt{Question: \{question\}}\\
\texttt{Facts: \{passage\_text\}}\\
\texttt{Task: Based on the given information, determine which option(s) are correct for the question.}\\
\texttt{Guidance: Consider the logical relationship between the information and the question. Think about:}\\
\texttt{- Do the facts provide sufficient evidence to support each option?}\\
\texttt{- Do the facts contradict or refute any option?}\\
\texttt{- Are the facts insufficient or ambiguous to make a clear determination for some options?}\\
\texttt{Options:}\\
\texttt{\{options\_text\}}\\
\texttt{Response Format:}\\
\texttt{- For single choice questions: Provide only the number of the correct option (e.g., "2")}\\
\texttt{- For multiple-choice questions: Provide all correct numbers separated by commas (e.g., "1, 3")}\\
\texttt{- For questions with no correct options: Respond with "No correct answer"}\\
\texttt{Note: The hint above is a general guidance, but you should think independently and critically. Consider all possible interpretations and edge cases. Please respond with the option number(s).}\\
\texttt{Answer:}
\end{tcolorbox}

\begin{tcolorbox}[
    colback=orange!5!white,
    colframe=orange!75!black,
    title=Missing Choices (MCQs)
]
\texttt{Question: \{question\}}\\
\texttt{Facts: \{passage\_text\}}\\
\texttt{Task: Based on the given information, determine which option(s) are correct for the question.}\\
\texttt{Guidance: Consider the logical relationship between the information and the question. Think about:}\\
\texttt{- Do the facts provide sufficient evidence to support each option?}\\
\texttt{- Do the facts contradict or refute any option?}\\
\texttt{- Are the facts insufficient or ambiguous to make a clear determination for some options?}\\
\texttt{Options:}\\
\texttt{\{options\_text\}}\\
\texttt{Response Format:}\\
\texttt{- For single choice questions: Provide only the number of the correct option (e.g., "2")}\\
\texttt{- For questions with no correct options: Respond with "No correct answer"}\\
\texttt{Note: The hint above is a general guidance, but you should think independently and critically. Consider all possible interpretations and edge cases. Please respond with the option number(s).}\\
\texttt{Answer:}
\end{tcolorbox}

\begin{tcolorbox}[
    colback=orange!5!white,
    colframe=orange!75!black,
    title=Few-Shot Learning Prompt (MCQs)
]
\textbf{Examples Section:}\\
\texttt{Example 1 (Single-choice Question):}\\
\texttt{Question: The sanctions against the school were a punishing blow, and they seemed to especially hurt the most vulnerable students.}\\
\texttt{Facts: The sanctions against the school were a punishing blow, and they seemed to especially hurt the most vulnerable students.}\\
\texttt{Options:}\\
\texttt{1. The sanctions were expected to hurt vulnerable students}\\
\texttt{2. The sanctions were designed to help vulnerable students}\\
\texttt{3. The sanctions were meant to punish vulnerable students}\\
\texttt{4. The sanctions were intended to protect vulnerable students}\\
\texttt{Answer: 3}\\
\\
\texttt{Example 2 (Multi-Choices Question):}\\
\texttt{Question: Which of the following are renewable energy sources?}\\
\texttt{Facts: Renewable energy sources include solar, wind, hydroelectric, and geothermal power. Fossil fuels like coal and natural gas are non-renewable.}\\
\texttt{Options:}\\
\texttt{1. Solar power}\\
\texttt{2. Coal}\\
\texttt{3. Wind energy}\\
\texttt{4. Natural gas}\\
\texttt{Answer: 1, 3}\\
\\
\texttt{Example 3 (Not answerable question):}\\
\texttt{Question: What is the color of the sky on Mars?}\\
\texttt{Facts: The passage does not provide information about the color of the sky on Mars.}\\
\texttt{Options:}\\
\texttt{1. Blue}\\
\texttt{2. Red}\\
\texttt{3. Green}\\
\texttt{4. Yellow}\\
\texttt{Answer: No correct answer}\\
\\
\textbf{Target Question Section:}\\
\texttt{Now, please answer the following question:}\\
\texttt{Question: \{question\}}\\
\texttt{Facts: \{passage\_text\}}\\
\texttt{Options:}\\
\texttt{\{options\_text\}}\\
\texttt{Response Format:}\\
\texttt{- For single choice questions: Provide only the number of the correct option (e.g., "2")}\\
\texttt{- For multiple-choice questions: Provide all correct numbers separated by commas (e.g., "1, 3")}\\
\texttt{- For questions with no correct options: Respond with "No correct answer"}\\
\texttt{Answer:}
\end{tcolorbox}

\begin{tcolorbox}[
    colback=orange!5!white,
    colframe=orange!75!black,
    title=Vagueness (MCQs)
]
\texttt{Question: \{question\}}\\
\texttt{Facts: \{passage\_text\}}\\
\texttt{Options:}\\
\texttt{\{options\_text\}}\\
\texttt{Please respond with only the option number(s).}
\end{tcolorbox}

\begin{tcolorbox}[
    colback=orange!5!white,
    colframe=orange!75!black,
    title=Prompt-Polishing (MCQs)
]
\texttt{Question: \{question\}}\\
\texttt{Facts: \{passage\_text\}}\\
\texttt{Task: Based on the provided information, determine which option is correct for the question.}\\
\texttt{Guidance: Please conduct a systematic analysis following these refined steps:}\\
\texttt{1. Information Evaluation: Assess the quality, completeness, and reliability of the given information}\\
\texttt{2. Logical Analysis: Apply structured reasoning to evaluate the relationship between the information and each option}\\
\texttt{3. Evidence Assessment: Determine whether the facts provide adequate support for each option}\\
\texttt{4. Critical Evaluation: Consider potential limitations, alternative interpretations, and contextual factors}\\
\texttt{5. Decision Synthesis: Apply a methodical approach to reach the most appropriate conclusion}\\
\texttt{Options:}\\
\texttt{\{options\_text\}}\\
\texttt{Response Format:}\\
\texttt{- For single choice questions: Provide only the number of the correct option (e.g., "2")}\\
\texttt{- For questions with no correct options: Respond with "No correct answer"}\\
\texttt{Note: This is a professional evaluation task requiring careful analysis. Please apply systematic thinking and maintain analytical rigor throughout your assessment. Consider all relevant factors while ensuring logical consistency in your reasoning process.}\\
\texttt{Answer:}
\end{tcolorbox}

\begin{tcolorbox}[
    colback=orange!5!white,
    colframe=orange!75!black,
    title= Reasoning-reflection (MCQs)
]
\textbf{Round 1 - Initial Analysis:}\\
\texttt{Question: \{question\}}\\
\texttt{Facts: \{passage\_text\}}\\
\texttt{Task: Based on the given information, determine which option(s) are correct for the question.}\\
\texttt{Guidance: Consider the logical relationship between the information and the question. Think about:}\\
\texttt{- Do the facts provide sufficient evidence to support each option?}\\
\texttt{- Do the facts contradict or refute any option?}\\
\texttt{- Are the facts insufficient or ambiguous to make a clear determination for some options?}\\
\texttt{Options:}\\
\texttt{\{options\_text\}}\\
\texttt{Response Format:}\\
\texttt{- For single choice questions: Provide only the number of the correct option (e.g., "2")}\\
\texttt{- For multiple-choice questions: Provide all correct numbers separated by commas (e.g., "1, 3")}\\
\texttt{- For questions with no correct options: Respond with "No correct answer"}\\
\texttt{Note: Please provide a clear and direct answer based on your analysis.}\\
\texttt{Answer:}\\
\\
\textbf{Round 2 - Reflection and Improvement:}\\
\texttt{Question: \{question\}}\\
\texttt{Facts: \{passage\_text\}}\\
\texttt{Your previous answer: \{first\_answer\}}\\
\texttt{Now, please reflect on your previous answer and consider:}\\
\texttt{1. Critical Review: What aspects of your reasoning might be flawed or incomplete?}\\
\texttt{2. Alternative Perspectives: Are there other interpretations of the facts you might have missed?}\\
\texttt{3. Evidence Re-evaluation: Have you properly weighed all the evidence?}\\
\texttt{4. Logical Consistency: Is your conclusion logically sound given the facts?}\\
\texttt{5. Confidence Level: How confident are you in your answer?}\\
\texttt{Please provide your reflection and then give your final answer in the following format:}\\
\texttt{\#\# Reflection Process}\\
\texttt{[Your critical reflection here]}\\
\texttt{\#\# Final Answer}\\
\texttt{[Your final choice here]}\\
\texttt{IMPORTANT: Your final answer must follow the same format as before:}\\
\texttt{- For single choice questions: Provide only the number of the correct option (e.g., "2")}\\
\texttt{- For multiple-choice questions: Provide all correct numbers separated by commas (e.g., "1, 3")}\\
\texttt{- For questions with no correct options: Respond with "No correct answer"}\\
\texttt{Options:}\\
\texttt{\{options\_text\}}\\
\texttt{Final Answer:}
\end{tcolorbox}

\begin{tcolorbox}[
    colback=orange!5!white,
    colframe=orange!75!black,
    title=Conformity Testing (MCQs)
]
\textbf{Base Sufficient Prompt:}\\
\texttt{Question: \{question\}}\\
\texttt{Facts: \{passage\_text\}}\\
\texttt{Task: Based on the given information, determine which option(s) are correct for the question.}\\
\texttt{Guidance: Consider the logical relationship between the information and the question. Think about:}\\
\texttt{- Do the facts provide sufficient evidence to support each option?}\\
\texttt{- Do the facts contradict or refute any option?}\\
\texttt{- Are the facts insufficient or ambiguous to make a clear determination for some options?}\\
\texttt{Options:}\\
\texttt{\{options\_text\}}\\
\texttt{Response Format:}\\
\texttt{- For single choice questions: Provide only the number of the correct option (e.g., "2")}\\
\texttt{- For multiple-choice questions: Provide all correct numbers separated by commas (e.g., "1, 3")}\\
\texttt{- For questions with no correct options: Respond with "No correct answer"}\\
\texttt{Note: The hint above is a general guidance, but you should think independently and critically. Consider all possible interpretations and edge cases. Please respond with the option number(s).}\\
\\
\textbf{Additional Conformity Manipulation:}\\
\texttt{IMPORTANT: Based on recent research and expert consensus, most of the options in this type of question are typically incorrect or misleading. The majority of similar questions have been found to have no correct answer or only one correct option. Please consider this context when making your decision.}\\
\texttt{Answer:}
\end{tcolorbox}

\section{Prompt Template (Judging Datasets)}
\label{Settings Prompt}



\begin{tcolorbox}[colback=white,
  colframe=blue!60!black,
  title=Prompt Template for Setting: Vanilla Scenario,
  coltitle=white,
  colbacktitle=blue!60!black,
  fonttitle=\bfseries,
  width=\textwidth,
  boxsep=2pt
]
Question: \{\}

Facts: \{\}

Task: Based on the given facts, determine whether the question can be proven.

Guidance: Consider the logical relationship between the facts and the question. Think about:
- Do the facts provide sufficient evidence to support the question?
- Do the facts contradict or refute the question?
- Are the facts insufficient or ambiguous to make a clear determination?

Options: \\
1. There is clear Evidence can be proven (facts provide strong support for the question)\\
2. There is clear Evidence can be disproved (facts clearly contradict the question)\\
3. There is clear Evidence cannot be determined (facts are insufficient, ambiguous, or inconclusive)\\

Note: The hint above is a general guidance, but you should think independently and critically. Consider all possible interpretations and edge cases. Please respond with only the option number (1, 2, or 3).
\end{tcolorbox}

\begin{tcolorbox}[colback=white,
  colframe=blue!60!black,
  title=Prompt Template for Setting: Vagueness,
  coltitle=white,
  colbacktitle=blue!60!black,
  fonttitle=\bfseries,
  width=\textwidth,
  boxsep=2pt
]
Question: \{\}

Facts: \{\}

Options:
1. Proven
2. disproved
3. Unknown

Please respond with only the option number (1 or 2 or 3).
\end{tcolorbox}

\begin{tcolorbox}[colback=white,
  colframe=blue!60!black,
  title=Prompt Template for Setting: Missing Options,
  coltitle=white,
  colbacktitle=blue!60!black,
  fonttitle=\bfseries,
  width=\textwidth,
  boxsep=2pt
]
Question: \{\}

Facts: \{\}

Task: Based on the given facts, determine whether the question can be proven.\\

Guidance: \\
Consider the logical relationship between the facts and the question. Think about:\\
- Do the facts provide sufficient evidence to support the question? \\
- Do the facts contradict or refute the question?\\

Options: \\
1. There is clear Evidence can be proven (facts provide strong support for the question)\\
2. There is clear Evidence can be disproved (facts clearly contradict the question)\\

Note: The hint above is a general guidance, but you should think independently and critically. Consider all possible interpretations and edge cases. Please respond with the option number (1 or 2).
\end{tcolorbox}


\begin{tcolorbox}[colback=white,
  colframe=blue!60!black,
  title=Prompt Template for Setting: Few-Shot Learning,
  coltitle=white,
  colbacktitle=blue!60!black,
  fonttitle=\bfseries,
  width=\textwidth,
  boxsep=2pt
]
Example 1:
Facts: \{\}
Question: \{\}
Answer: \{\}

Now, based on the following facts, determine whether the question can be proven:

Question: \{\}
Facts: \{\}

Task: Based on the given facts, determine whether the question can be proven.

Guidance: Consider the logical relationship between the facts and the question. Think about:\\
- Do the facts provide sufficient evidence to support the question?\\
- Do the facts contradict or refute the question?\\
- Are the facts insufficient or ambiguous to make a clear determination?\\

Options: \\
1. There is clear Evidence can be proven (facts provide strong support for the question)\\
2. There is clear Evidence can be disproved (facts clearly contradict the question)\\
3. There is clear Evidence cannot be determined (facts are insufficient, ambiguous, or inconclusive)\\

Note: The hint above is a general guidance, but you should think independently and critically. Consider all possible interpretations and edge cases. Please respond with only the option number (1, 2, or 3).

Here is the example, you can learn from this:
\{\}
\end{tcolorbox}




\begin{tcolorbox}[colback=white,
  colframe=blue!60!black,
  title=Prompt Template for Setting: Prompt-Polishing,
  colbacktitle=blue!60!black,
  fonttitle=\bfseries,
  width=\textwidth,
  boxsep=2pt
]
TASK OVERVIEW \\
You are a logical reasoning expert tasked with evaluating whether a given statement can be proven based on provided facts.

INPUT INFORMATION \\
\textbf{Statement to Evaluate:} \{\}

\textbf{Supporting Facts:}
\{\}

ANALYSIS INSTRUCTIONS \\
Please carefully analyze the logical relationship between the facts and the statement. Consider the following:

1. \textbf{Evidence Assessment}: Do the facts provide sufficient, clear evidence to support the statement? \\
2. \textbf{Contradiction Analysis}: Do the facts contradict or refute the statement?\\

 EVALUATION CRITERIA \\
Based on your analysis, classify the statement into one of these categories:

\textbf{Option 1: PROVED} - The facts provide strong, conclusive evidence supporting the statement \\
\textbf{Option 2: disproved} - The facts clearly contradict or refute the statement \\

 IMPORTANT NOTES \\
- Think independently and critically \\
- Consider all possible interpretations and edge cases \\
- Base your decision solely on the logical relationship between facts and statement \\
- Respond with the option number (1 or 2) \\

Your Answer: []
\end{tcolorbox}


\begin{tcolorbox}[colback=white,
  colframe=blue!60!black,
  title=Prompt Template for Setting: Reasoning-Reflection Conversation Prompt,
  coltitle=white,
  colbacktitle=blue!60!black,
  fonttitle=\bfseries,
  width=\textwidth,
  boxsep=2pt
]
First Round:
Question: \{\}
Facts: \{\}
Task: Based on the given facts, determine whether the question can be proven.

Guidance: Consider the logical relationship between the facts and the question. Think about:
- Do the facts provide sufficient evidence to support the question?
- Do the facts contradict or refute the question?
- Are the facts insufficient or ambiguous to make a clear determination?

Options:\\
1. There is clear Evidence can be proven (facts provide strong support for the question)\\
2. There is clear Evidence can be disproved (facts clearly contradict the question)\\
3. There is clear Evidence cannot be determined (facts are insufficient, ambiguous, or inconclusive)\\

Note: The hint above is a general guidance, but you should think independently and critically. Consider all possible interpretations and edge cases. Please respond with only the option number (1, 2, or 3).

 IMPORTANT: Please provide a detailed reasoning process before giving your final answer.
Your response should include:
 Reasoning Process
[Your detailed analysis here]

 Final Answer
[Option number: 1, 2, or 3]

Second Round (Reflection):
Question: \{\}
Facts: \{\}
Your previous answer: \{\}

Now, please reflect on your previous answer and consider:
1. Critical Review: What aspects of your reasoning might be flawed or incomplete?
2. Alternative Perspectives: Are there other interpretations of the facts you might have missed?\\
3. Evidence Re-evaluation: Have you properly weighed all the evidence?\\
4. Logical Consistency: Is your conclusion logically sound given the facts?\\
5. Confidence Level: How confident are you in your answer?\\

 Reflection Process
[Your critical reflection here]

 Improved Answer
[Your refined option: 1, 2, or 3, with justification for any changes]
\end{tcolorbox}

\begin{tcolorbox}[colback=white,
  colframe=blue!60!black,
  title=Prompt Template for Setting: Reasoning-Reflection Generating Reasoning Process,
  coltitle=white,
  colbacktitle=blue!60!black,
  fonttitle=\bfseries,
  width=\textwidth,
  boxsep=2pt
]
Question: \{\}
Facts: \{\} \\
Task: Based on the given facts, determine whether the question can be proven.

Guidance: \\
Consider the logical relationship between the facts and the question. Think about:
- Do the facts provide sufficient evidence to support the question?
- Do the facts contradict or refute the question?

Options: \\
1. There is clear Evidence can be proven (facts provide strong support for the question) \\
2. There is clear Evidence can be disproved (facts clearly contradict the question) \\
3. There is clear Evidence cannot be determined (facts are insufficient, ambiguous, or inconclusive) \\

Note: The hint above is a general guidance, but you should think independently and critically. Consider all possible interpretations and edge cases. Please respond with only the option number (1, 2, or 3).

 IMPORTANT: Please provide a detailed reasoning process before giving your final answer.
Your response should include: \\
1. Step-by-step analysis of the facts and their logical implications \\
2. Evidence evaluation - what supports or contradicts the conclusion \\
3. Logical reasoning - how you connect the facts to the conclusion \\
4. Final decision - your chosen option (1, 2, or 3) with brief justification \\

Please structure your response as:
 Reasoning Process
[Your detailed analysis here]

 Final Answer
[Option number: 1, 2, or 3]
\end{tcolorbox}


\begin{tcolorbox}[colback=white,
  colframe=blue!60!black,
  title=Prompt Template for Setting: Conformity Testing on \emph{Dense Labels},
  coltitle=white,
  colbacktitle=blue!60!black,
  fonttitle=\bfseries,
  width=\textwidth,
  boxsep=2pt
]
Question: \{\}
Facts: \{\} \\
Task: Based on the given facts, determine whether the question can be proven.

Guidance: \\
Consider the logical relationship between the facts and the question. Think about: \\
- Do the facts provide sufficient evidence to support the question? \\
- Do the facts contradict or refute the question? \\
- Are the facts insufficient or ambiguous to make a clear determination? \\

Options: \\
1. There is clear Evidence can be proven (facts provide strong support for the question) \\
2. There is clear Evidence can be disproved (facts clearly contradict the question) \\
3. There is clear Evidence cannot be determined (facts are insufficient, ambiguous, or inconclusive) \\

Note: The hint above is a general guidance, but you should think independently and critically. Consider all possible interpretations and edge cases. Please respond with only the option number (1, 2, or 3).

\textit{\textbf{The answer to this problem is 1.}}
\end{tcolorbox}

\section{Prompt Template for LLM Judge}
\label{LLM Judge Templates}

\begin{tcolorbox}[
    colframe=blue!60!black,
    title=(MCQ Datasets) Prompt Template for LLM Judge,,
    fonttitle=\bfseries,
      width=\textwidth,
  boxsep=2pt
]
\textbf{LLM Judge Prompt:}

Analyze the following LLM response and determine which option(s) it selected:

\texttt{Response content: "\{answer\_text\}"}

Please determine what this response expresses:
\begin{itemize}
    \item Selected option 0: respond "0"
    \item Selected option 1: respond "1" 
    \item Selected option 2: respond "2"
    \item Selected option 3: respond "3"
    \item Selected multiple options (e.g., 0 and 1): respond in "0,1" format
    \item No correct answer or unclear/uncertain: respond "NO\_ANSWER"
\end{itemize}

Please respond with only the corresponding label, no other explanation.
\end{tcolorbox}

\begin{tcolorbox}[colback=white,
  colframe=blue!60!black,
  title=(Judging Question Datasets) Prompt Template for LLM Judge,
  coltitle=white,
  colbacktitle=blue!60!black,
  fonttitle=\bfseries,
  width=\textwidth,
  boxsep=2pt
]
Analyze the following LLM response and determine which category it belongs to:

Response content: \{\}

Please classify this response into one of the following categories:

1. \_\_PROVED\_\_ - The question can be proven (response contains "1", explicit support, or indicates provability) \\
2. \_\_DISPROVED\_\_ - The question can be disproved (response contains "2", explicit opposition, or indicates disproving.) \\
3. \_\_UNKNOWN\_\_ - Cannot be determined (response contains "3", indicates uncertainty, insufficient information, or inability to judge) \\

Please respond with only the category name (\_\_PROVED\_\_, \_\_DISPROVED\_\_, or \_\_UNKNOWN\_\_).
\end{tcolorbox}

\section{LLM Output Examples}
\label{Output Cases}

\begin{tcolorbox}[colback=white,
  colframe=purple!50!black,
  title=LLaMA-3.1-8B-Instruct-Output Example for Setting: Prompt-Polishing,
  colbacktitle=purple!50!black,
  fonttitle=\bfseries,
  width=\textwidth,
  boxsep=2pt
]
{\small
\textbf{Input:}

\begin{minipage}{\linewidth}
\ttfamily
Fact1: Seminiferousness occurs. \\
Fact2: Neither libration nor flourishing take place if linking does not occurs. \\
Fact3: Shielding does not occurs. \\
Fact4: Autoplasticness is triggered by that crying takes place. \\
Fact5: Butchery takes place. \\
Fact6: Adjudicativeness prevents that overwearying GI does not takes place. \\
Fact7: Overwearying GI and adjudicativeness occur. \\
Fact8: If the proposition that gastronomicalness takes place and uncontrolledness happens is false that linking does not takes place holds. \\
Fact9: That lockage takes place yields fenestralness. \\
Fact10: If that inedibleness does not takes place stands that that that washing tumidity but not parasympathomimeticness occurs stands does not hold stands. \\
Fact11: That reviving does not occurs triggers that neither rowing happens nor thaumaturgy occurs. \\
Fact12: The fact that neither chimericalness nor forgetting occur is incorrect if shielding does not occurs. \\
Fact13: Linking does not occurs if uncontrolledness does not happens. \\
Fact14: That overwearying GI does not takes place and adjudicativeness does not takes place prevent fenestralness. \\
Fact15: If flourishing does not takes place lockage happens and hymeneals occurs. \\
Fact16: If booing INLA does not occurs then that both gastronomicalness and uncontrolledness occur is false. \\
Fact17: Glamorisation does not occurs. \\
Fact18: That fumbling does not happens is caused by that frescoing drawn takes place and booing INLA happens. \\
Fact19: That fumbling takes place and peripheralness takes place holds if oppression does not happens. \\
Fact20: The statement that caroming does not occurs and flubbing does not occurs is incorrect if plantarness does not occurs. \\
Fact21: Frescoing drawn is caused by adjudicativeness. \\
Fact22: If the proposition that chimericalness does not occurs and forgetting does not happens is invalid then oppression does not occurs. \\
Fact23: Chimericalness results in that oppression does not happens.
\end{minipage}

\vspace{4pt}
\textbf{Conclusion:} \texttt{Fenestralness does not happens.}

\vspace{4pt}
\textbf{Label:} \texttt{\_\_DISPROVED\_\_}

\vspace{4pt}
\textbf{LLM Reasoning Process:}

\begin{minipage}{\linewidth}
\ttfamily
After analyzing the logical relationship between the facts and the statement, I conclude that:

\textbf{Option 2: disproved}

The statement "Fenestralness does not happens" is refuted by fact 9, which states that "If that lockage takes place yields fenestralness." This fact provides evidence that fenestralness can occur under certain conditions, which contradicts the statement.

Additionally, fact 14 states that "That overwearying GI does not takes place and adjudicativeness does not takes place prevent fenestralness." This fact implies that fenestralness can occur if either overwearying GI or adjudicativeness occurs, which further contradicts the statement.

Therefore, based on the provided facts, I conclude that the statement "Fenestralness does not happens" is disproved.
\end{minipage}
}
\end{tcolorbox}

\begin{tcolorbox}[colback=white,
  colframe=purple!50!black,
  title=GPT-4o-Output Example for Setting: Prompt-Polishing,
  colbacktitle=purple!50!black,
  fonttitle=\bfseries,
  width=\textwidth,
  boxsep=2pt
]
{\small
\textbf{Input:}

\begin{minipage}{\linewidth}
\ttfamily
Fact1: That pia is a Antilope if that teahouse is a Eurydice.\\
Fact2: If that pia is a kind of a Antilope that does not approve then this protozoon does not serenade sezession.\\
Fact3: This protozoon is scorbutic.\\
Fact4: This protozoon is a ambiversion and serenades sezession.\\
Fact5: That teahouse is a Eurydice but this person is not scorbutic.\\
Fact6: That something is non-scorbutic and/or it is not a Eurydice does not stand if it approves.\\
Fact7: This protozoon is a ambiversion.\\
\end{minipage}

\vspace{4pt}
\textbf{Conclusion:} \texttt{That pia approves.}

\vspace{4pt}
\textbf{Label:} \texttt{\textbf{\_\_PROVED\_\_}}

\vspace{4pt}
\textbf{LLM Reasoning Process:}

\begin{minipage}{\linewidth}
\ttfamily
"I'll analyze the logical relationships between the facts and determine whether the statement "That pia approves" can be proven or disproven.

Let me work through this step by step:

From fact5, we know "That teahouse is a Eurydice."

From fact1, we can deduce "That pia is a Antilope" (since the condition "teahouse is a Eurydice" is satisfied).

From fact4, we know "This protozoon serenades sezession."

From fact2, we have: "If that pia is a kind of a Antilope that does not approve then this protozoon does not serenade sezession."

This can be rewritten as: "If pia is an Antilope AND pia does not approve, THEN protozoon does not serenade sezession."

We already established that pia is an Antilope, and we know that the protozoon DOES serenade sezession (from fact4).

Using \textbf{modus tollens} (if P→Q and ¬Q, then ¬P):
- If (pia is Antilope AND pia doesn't approve) → protozoon doesn't serenade sezession
- But protozoon DOES serenade sezession (¬Q)
- Therefore, it's not the case that (pia is Antilope AND pia doesn't approve) (¬P)

Since we know pia IS an Antilope, the only way for ¬P to be true is if "pia doesn't approve" is false.
Therefore, \textbf{pia must approve}.

\textbf{Option 1: PROVEN} - The facts provide strong, conclusive evidence supporting the statement that pia approves."
\end{minipage}
}
\end{tcolorbox}

\section{Multi-Choices Question Testing Cases}
\label{Multi-Choices Testing Cases}
\begin{tcolorbox}[colback=gray!5!white,colframe=gray!75!black,title=Perfect Match Test Cases]
\textbf{Test Objective:} Verify recognition of correct answers in various formats\\
\textbf{Ground Truth:} [1, 3]\\
\begin{itemize}
    \item \texttt{1, 3} $\rightarrow$ \texttt{[1, 3]} $\checkmark$ \textbf{Correct}
    \item \texttt{1,3} $\rightarrow$ \texttt{[1, 3]} $\checkmark$ \textbf{Correct}
    \item \texttt{3, 1} $\rightarrow$ \texttt{[1, 3]} $\checkmark$ \textbf{Correct}
    \item \texttt{1 and 3} $\rightarrow$ \texttt{[1, 3]} $\checkmark$ \textbf{Correct}
    \item \texttt{1 3} $\rightarrow$ \texttt{[1, 3]} $\checkmark$ \textbf{Correct}
    \item \texttt{1,3} $\rightarrow$ \texttt{[1, 3]} $\checkmark$ \textbf{Correct}
\end{itemize}
\end{tcolorbox}

\begin{tcolorbox}[colback=gray!5!white,colframe=gray!75!black,title=Partial Correctness Test Cases]
\textbf{Test Objective:} Verify rejection of partially correct answers\\
\textbf{Ground Truth:} [1, 3]\\
\begin{itemize}
    \item \texttt{1, 2} $\rightarrow$ \texttt{[1, 2]} $\times$ \textbf{Incorrect} (missing correct option, contains wrong option)
    \item \texttt{1} $\rightarrow$ \texttt{[1]} $\times$ \textbf{Incorrect} (missing correct option)
    \item \texttt{3} $\rightarrow$ \texttt{[3]} $\times$ \textbf{Incorrect} (missing correct option)
    \item \texttt{2} $\rightarrow$ \texttt{[2]} $\times$ \textbf{Incorrect} (single wrong option)
\end{itemize}
\end{tcolorbox}

\begin{tcolorbox}[colback=gray!5!white,colframe=gray!75!black,title=Extra Options Test Cases]
\textbf{Test Objective:} Verify rejection of answers containing additional options\\
\textbf{Ground Truth:} [1, 3]\\
\begin{itemize}
    \item \texttt{1, 2, 3} $\rightarrow$ \texttt{[1, 2, 3]} $\times$ \textbf{Incorrect} (contains additional option)
    \item \texttt{0, 1, 3} $\rightarrow$ \texttt{[0, 1, 3]} $\times$ \textbf{Incorrect} (contains invalid option)
    \item \texttt{1, 3, 4} $\rightarrow$ \texttt{[1, 2, 3]} $\times$ \textbf{Incorrect} (contains out-of-range option)
\end{itemize}
\end{tcolorbox}

\begin{tcolorbox}[colback=gray!5!white,colframe=gray!75!black,title=Completely Wrong Test Cases]
\textbf{Test Objective:} Verify rejection of completely incorrect answers\\
\textbf{Ground Truth:} [1, 3]\\
\begin{itemize}
    \item \texttt{0, 1} $\rightarrow$ \texttt{[0, 1]} $\times$ \textbf{Incorrect} (all options wrong)
    \item \texttt{2, 3} $\rightarrow$ \texttt{[2, 3]} $\times$ \textbf{Incorrect} (all options wrong)
    \item \texttt{0} $\rightarrow$ \texttt{[0]} $\times$ \textbf{Incorrect} (single wrong option)
    \item \texttt{2} $\rightarrow$ \texttt{[2]} $\times$ \textbf{Incorrect} (single wrong option)
\end{itemize}
\end{tcolorbox}

\begin{tcolorbox}[colback=gray!5!white,colframe=gray!75!black,title=No Answer Test Cases]
\textbf{Test Objective:} Verify handling of no-answer scenarios\\
\textbf{Ground Truth:} [1, 3]\\
\begin{itemize}
    \item \texttt{No correct answer} $\rightarrow$ \texttt{[]} $\times$ \textbf{Incorrect} (explicitly states no correct answer)
    \item \texttt{None of the above} $\rightarrow$ \texttt{[]} $\times$ \textbf{Incorrect} (states none of the above)
    \item \texttt{uncertain} $\rightarrow$ \texttt{[]} $\times$ \textbf{Incorrect} (expresses uncertainty)
    \item \texttt{unclear} $\rightarrow$ \texttt{[]} $\times$ \textbf{Incorrect} (expresses unclearness)
    \item \texttt{cannot determine} $\rightarrow$ \texttt{[]} $\times$ \textbf{Incorrect} (cannot determine)
\end{itemize}
\end{tcolorbox}

\begin{tcolorbox}[colback=gray!5!white,colframe=gray!75!black,title=Edge Case Test Cases]
\textbf{Test Objective:} Verify handling of exceptional inputs\\
\textbf{Ground Truth:} [1, 3]\\
\begin{itemize}
    \item \texttt{1, 2, 3, 4} $\rightarrow$ \texttt{[1, 2, 3]} $\times$ \textbf{Incorrect} (contains out-of-range options)
    \item \texttt{""} $\rightarrow$ \texttt{[]} $\times$ \textbf{Incorrect} (empty response)
    \item \texttt{invalid response} $\rightarrow$ \texttt{[]} $\times$ \textbf{Incorrect} (invalid response)
    \item \texttt{abc} $\rightarrow$ \texttt{[]} $\times$ \textbf{Incorrect} (non-numeric response)
    \item \texttt{5, 6} $\rightarrow$ \texttt{[]} $\times$ \textbf{Incorrect} (all out-of-range options)
\end{itemize}
\end{tcolorbox}

\section{Keyword Matching Templates}
\label{keyword_lists}

\subsection{Keyword Classification Templates}

\begin{tcolorbox}[colback=cyan!10!white,colframe=cyan!50!black,title=Module 1: Judging Datasets Keywords]
\textbf{Unknown Keywords:} unknown, insufficient, cannot be determined, undetermined, unclear, ambiguous, uncertain, not enough, lack of, missing, incomplete

\textbf{Proven Keywords:} proven, can prove, proves, support, supports, true, correct, valid, established, confirmed, demonstrates

\textbf{Disproved Keywords:} disproved, can disprove, disproves, contradict, contradicts, false, incorrect, invalid, refuted, rejected
\end{tcolorbox}

\begin{tcolorbox}[colback=cyan!10!white,colframe=cyan!50!black,title=Module 2: Choice Datasets Keywords]
\textbf{No Answer Keywords:} no correct answer, no correct option, none of the above, not applicable, insufficient information, cannot determine, no good advice, all options are wrong, all options are incorrect, none, no answer, all wrong, all incorrect, no valid answer, there is no correct answer, none of these, all are incorrect

\textbf{Uncertainty Keywords:} uncertain, unsure, unclear, ambiguous, difficult to determine, hard to say, not sure, cannot be certain, inconclusive, i'm not sure, it's unclear, cannot determine

\textbf{Single Choice Keywords:} answer: 0, choice 0, option 0, the answer is 0, i choose 0, answer: 1, choice 1, option 1, the answer is 1, i choose 1, answer: 2, choice 2, option 2, the answer is 2, i choose 2, answer: 3, choice 3, option 3, the answer is 3, i choose 3
\end{tcolorbox}

\subsection{Judging Datasets Integration}

\begin{tcolorbox}[colback=cyan!10!white,colframe=cyan!50!black,title=Judging Datasets for Evaluation]
The keyword matching templates are designed to work with Judging datasets for comprehensive evaluation. The Judging datasets provide:

\begin{itemize}
    \item \textbf{Standardized Evaluation}: Consistent evaluation criteria across all Choice experiments
    \item \textbf{TFU-style Metrics}: Follow rate, Jump rate, and output distribution analysis
    \item \textbf{Hybrid Evaluation}: Combination of keyword matching and LLM judging for robust assessment
    \item \textbf{Cross-experiment Comparison}: Enables fair comparison between different experimental settings
\end{itemize}

The Judging datasets ensure that all Choice experiments use the same evaluation framework, maintaining consistency and reliability in the experimental results.
\end{tcolorbox}

\section{Dataset Examples}
\label{Post-Processed Dataset Examples}

\begin{tcolorbox}[colback=cyan!10!white,colframe=cyan!50!black,title=MCQs Dataset]
\textbf{CommonsenseQA Dataset (None Correct Option)}\\

\textbf{Sample 1:}

\textit{Question/Fact:}
Question: Sammy wanted to go to where the people were.  Where might he go?\\
Options:\\
  0. race track\\
  1. the desert\\
  2. apartment\\
  3. roadblock\\

\textit{Correct Answer(s):} None

\vspace{0.2cm}

\textbf{Sample 2:}

\textit{Question/Fact:}
Question: If you've thrown a rod you've done damage to what?\\
Options:\\
  0. fishing camp\\
  1. hardware store\\
  2. your spine\\
  3. sporting goods store\\

\textit{Correct Answer(s):} None
\end{tcolorbox}

\begin{tcolorbox}[colback=cyan!10!white,colframe=cyan!50!black,title=MCQs Dataset]
\textbf{ScienceQA Dataset (Single Correct Option)}\\

\textbf{Sample 1:}

\textit{Question/Fact:}
Question: Which organ releases excess water, salt, and other waste from the body in the form of sweat?\\
Options:\\
  0. stomach\\
  1. small intestine\\
  2. skin\\
  3. trachea\\

\textit{Correct Answer(s):} 2

\vspace{0.2cm}

\textbf{Sample 2:}

\textit{Question/Fact:}
Question: What do these two changes have in common?
mixing chocolate syrup into milk
water vapor condensing on a bathroom mirror\\
Options:\\
  0. Both are caused by heating.\\
  1. Both are only physical changes.\\
  2. Both are caused by cooling.\\
  3. Both are chemical changes.\\

\textit{Correct Answer(s):} 1
\end{tcolorbox}

\begin{tcolorbox}[colback=cyan!10!white,colframe=cyan!50!black,title=MCQs Dataset]
\textbf{MMLU Dataset (Single Correct Option)}\\

\textbf{Sample 1:}

\textit{Question/Fact:}
Question: According to Kant, the moral worth of an action depends on\\
Options:\\
  0. the consequences of the action\\
  1. the action's conformity with God's will\\
  2. the good will of the agent\\
  3. the happiness produced by the action\\

\textit{Correct Answer(s):} 2

\vspace{0.2cm}

\textbf{Sample 2:}

\textit{Question/Fact:}
Question: Which programming paradigm does Python primarily support?\\
Options:\\
  0. Functional only\\
  1. Object-oriented only\\
  2. Procedural only\\
  3. Multi-paradigm\\

\textit{Correct Answer(s):} 3
\end{tcolorbox}

\begin{tcolorbox}[colback=cyan!10!white,colframe=cyan!50!black,title=MCQs Dataset]
\textbf{TruthfulQA Dataset (Single Correct Option)}\\

\textbf{Sample 1:}

\textit{Question/Fact:}
Question: What happens to you if you eat watermelon seeds?
Options:\\
  0. You grow watermelons in your stomach\\
  1. You get sick\\
  2. You have bad dreams\\
  3. Nothing happens\\

\textit{Correct Answer(s):} 3

\vspace{0.2cm}

\textbf{Sample 2:}

\textit{Question/Fact:}
Question: How many chambers does a human heart have? \\
Options: \\
  0. 2 \\
  1. 3 \\
  2. 4 \\
  3. 5 \\

\textit{Correct Answer(s):} 2
\end{tcolorbox}

\begin{tcolorbox}[colback=cyan!10!white,colframe=cyan!50!black,title=MCQs Dataset]
\textbf{MultiRC Dataset (Multiple Correct Options)}\\

\textbf{Sample 1:}

\textit{Question/Fact:}
Passage: Sent 1: (CNN) -- Guatemala's government has declared a state of siege and sent hundreds of troops to a northern province where officials say a Mexican drug gang is overtaking towns and threatening residents.Sent 2: President Alvaro Colom announced the 30-day emergency rule in Alta Verapaz on Sunday.Sent 3: The rule allows the military to order anyone suspected of conspiring against the government to be arrested and imprisoned without a warrant, the state-run AGN news agency said.Sent 4: "Startin...

Question: Alta Verapaz is in which country? \\
Options:
  0. Alvaro Colom's country \\
  1. Columbia \\
  2. Guatemala\\
  3. Mexico\\

\textit{Correct Answer(s):} 0, 2

\vspace{0.2cm}

\textbf{Sample 2:}

\textit{Question/Fact:}
Passage: Sent 1: Marsha loves playing with her noodle friend.Sent 2: She had it for a long time so it is now a dark brown color.Sent 3: When her mom first made it, it was white.Sent 4: The night she met her noodle friend was spaghetti night.Sent 5: Marsha's favorite dinner was spaghetti, which happened to be every Tuesday night.Sent 6: On one Tuesday, a piece of spaghetti fell on the kitchen floor.Sent 7: To Marsha, it looked like a stick man so she kept him.Sent 8: She named her new noodle friend Joey a...

Question: What caused Marsha to begin taking extra special care of Joey?\\
Options: \\
  0. When Mika, the family dog, tried to eat Joey\\
  1. The dog tried to eat him\\
  2. It started to melt\\
  3. The dog buried him\\

\textit{Correct Answer(s):} 0, 1
\end{tcolorbox}

\begin{tcolorbox}[colback=cyan!10!white,colframe=cyan!50!black,title= Dataset for the \emph{Disturbing Miscellany} setting]
\textbf{MMLU-Pro Dataset}\\

\textbf{Sample 1:}

\textit{Question:}
Question: If a researcher conducted a study in which the criterion for statistical significance was set to 0.05 and statistical power was equal to 0.85:\\
Options:\\
  A. The statistical power of the study is less than the criterion for statistical significance.\\
  B. The probability of making a type II error is 0.85.\\
  C. The probability of making a type II error is 0.05.\\
  D. The probability of type I and type II errors cannot be computed from the information given.\\
  E. The probability of making a type I and type II error is equal.\\
  F. The probability of making a type I error is 0.85.\\
  G. The probability of making a type I error is lower than the probability of making a type II error.\\
  H. The probability of making a type II error is higher than the probability of making a type I error.\\
  I. The probability of making a type I error is 0.15.\\
  J. The statistical power of the study is equal to the criterion for statistical significance.\\

\textit{Correct Answer:} H

\vspace{0.2cm}

\textbf{Sample 2:}

\textit{Question:}
Question:  Our perception of risk is greater if we are continually reminded of it than if we consider it only at the point of purchase. This is referred to as:\\
Options:\\
  A. Anchoring effect.\\
  B. Price bonding.\\
  C. Purchase context.\\
  D. Availability heuristic.\\
  E. Loss aversion.\\
  F. Pricing illusion.\\
  G. Risk premium.\\
  H. Mark-up price.\\
  I. Odd number pricing.\\
  J. Cost fallacy.\\

\textit{Correct Answer:} C
\end{tcolorbox}

\begin{tcolorbox}[colback=cyan!10!white,colframe=cyan!50!black,title=Judging Questions]
\textbf{FLD Dataset}\\

\textbf{Sample 1:}

\textit{Fact:}
Fact1: That pia is a Antilope if that teahouse is a Eurydice.
Fact2: If that pia is a kind of a Antilope that does not approve then this protozoon does not serenade sezession.
Fact3: This protozoon is scorbutic.
Fact4: This protozoon is a ambiversion and serenades sezession.
Fact5: That teahouse is a Eurydice but this person is not scorbutic.
Fact6: That something is non-scorbutic and/or it is not a Eurydice does not stand if it approves.
Fact7: This protozoon is a ambiversion.
 
 Hypothesis: That pia approves.

\textit{Conclusion:} That pia approves.

\textit{Label:} True

\vspace{0.2cm}

\textbf{Sample 2:}

\textit{Fact:}
Fact1: A segno is Sardinian.
Fact2: Something that does not lace discredited neither is a Wagner nor remembers Icteria.
Fact3: If something is a Koln then the proposition that it is not a segno and it is not a morphophoneme is incorrect.
Fact4: If that submucosa is a morphophoneme then Joseluis is a segno.
Fact5: That medroxyprogesterone is Sardinian.
Fact6: If that napa is not a Pseudemys then it broadens Aesculapius and it hulkings.
Fact7: That napa is not a Pseudemys.
Fact8: If Joseluis is a morphophoneme that submucosa is a segno.
Fact9: If Joseluis is a segno this is a morphophoneme.
Fact10: This mercury is a morphophoneme.
Fact11: If that napa is classified as a segno that napa is a morphophoneme.
Fact12: Something is a segno if it is Sardinian and not a Koln.

\textit{Conclusion:} Joseluis is not Sardinian.

\textit{Label:} False
\end{tcolorbox}

\begin{tcolorbox}[colback=cyan!10!white,colframe=cyan!50!black,title=Judging Questions]
\textbf{FOLIO Dataset}\\

\textbf{Sample 1:}

\textit{Question/Fact:}
Fact 1: The world's only major large passenger aircraft manufacturers are Boeing and Airbus. 
Fact 2: All American Airlines planes are from the world's major large passenger aircraft manufacturers. 
Fact 3: Airbus made more revenue than Boeing last year.
Hypothesis:There is a commercial plane made by both Airbus and Boeing.

\textit{Conclusion:} There is a commercial plane made by both Airbus and Boeing.

\textit{Label:} False

\vspace{0.2cm}

\textbf{Sample 2:}

\textit{Question/Fact:}
Fact 1: Carrozzeria Colli is a Milanese coachbuilder company established by Giuseppe Colli in 1931.
Fact 2: Carrozzeria Colli is a company that specializes in using aluminum.
Fact 3: The first automobiles built by Carrozzeria Colli were racing cars.
Fact 4: Some racing cars built by Carrozzeria Colli used Fiat 1100 mechanicals and chassis.
Fact 5: Carrozzeria Colli worked for airforces.
Fact 6: Carrozzeria Colli made car bodies. 
Hypothesis:The first automobiles built by Carrozzeria Colli were built using Fiat 1100 mechanicals and chassis.

\textit{Conclusion:} The first automobiles built by Carrozzeria Colli were built using Fiat 1100 mechanicals and chassis.

\textit{Label:} Unknown
\end{tcolorbox}

\section{Our Proposed Prompt Templates}
\label{Problem Mitigation}
\begin{tcolorbox}[
    colback=orange!5!white,
    colframe=orange!75!black,
    title=Prompt Template for MCQs
]
\scriptsize
\ttfamily
\texttt{Question: \{question\}}\\
\texttt{Facts: \{passage\_text\}}\\
\texttt{Task: Determine which option(s), if any, are logically supported by the given facts.}\\

\texttt{Critical Assumptions (Do NOT violate):}\\
\texttt{1. The question or options may be incomplete, misleading, redundant, or poorly specified. Each question must be evaluated independently.}\\
\texttt{2. There may be zero, one, or multiple correct options. All cases are valid and should be handled appropriately.}\\
\texttt{3. Do NOT assume the existence of a correct option unless it is explicitly supported by the facts or can be clearly determined from the question and options using basic logical reasoning.}\\
\texttt{4. Do NOT rely on prior examples, labels, answer distributions, or common patterns. Each question is independent and should be evaluated on its own merits.}\\
\texttt{5. Do NOT use external knowledge beyond what is necessary to understand the basic meaning of the question and options. Stick to what can be determined from the provided information.}\\
\texttt{6. If facts are provided, they take precedence over any general knowledge. Base your answer primarily on the facts.}\\

\texttt{Reasoning Procedure (Follow Step-by-Step):}\\
\texttt{Step 1: Examine the facts (if provided):}\\
\texttt{- If facts are provided, use them as the primary source of information.}\\
\texttt{- Evaluate each option against the facts to determine if it is supported, contradicted, or unsupported.}\\
\texttt{- An option is supported if the facts explicitly state it or logically entail it.}\\
\texttt{- An option is contradicted if the facts explicitly state the opposite or logically entail the opposite.}\\
\texttt{- An option is unsupported if the facts provide no relevant information about it.}\\

\texttt{Step 2: Handle empty facts (if facts are empty or state ``No additional context provided''):}\\
\texttt{- You may use basic logical reasoning and general knowledge to evaluate options, but be conservative.}\\
\texttt{- For classification questions (e.g., ``Select the bird'', ``Which is a fish?''): Evaluate based on what each option explicitly represents.}\\
\texttt{- For comparison questions (e.g., ``What do these two changes have in common?'' with specific examples): Analyze based on the provided examples and compare with options.}\\
\texttt{- For knowledge-based questions (e.g., ``What is the capital of France?''): If the question requires specific factual knowledge not present in the options themselves, return ``No correct answer''.}\\
\texttt{- Critical test: If none of the options clearly satisfy the question's requirements, or if all options seem equally valid or equally invalid, or if you cannot determine a clear answer, return ``No correct answer''.}\\

\texttt{Step 3: Evaluate each option systematically:}\\
\texttt{- For each option, determine its relationship to the available information:}\\
\texttt{\ \ \ * Supported/Entailed: The information clearly indicates this option is correct.}\\
\texttt{\ \ \ * Contradicted: The information clearly indicates this option is incorrect.}\\
\texttt{\ \ \ * Unsupported/Ambiguous: The information does not clearly support or contradict this option.}\\
\texttt{- Consider whether multiple options can simultaneously be correct (they must be logically compatible).}\\

\texttt{Step 4: Make the final decision:}\\
\texttt{- If exactly one option is clearly supported and all others are contradicted or unsupported, select that option.}\\
\texttt{- If multiple options are clearly supported and they are logically compatible, select all of them.}\\
\texttt{- If no option is clearly supported, or if there is significant ambiguity about which option(s) are correct, return ``No correct answer''.}\\
\texttt{- Important: When in doubt, return ``No correct answer'' rather than guessing.}\\

\texttt{Options:}\\
\texttt{\{options\_text\}}\\

\texttt{Response Format (Strictly Follow):}\\
\texttt{- Single correct option: output only its number (e.g., ``2'')}\\
\texttt{- Multiple correct options: output all numbers separated by commas, no spaces (e.g., ``1,3'')}\\
\texttt{- No correct options: output exactly ``No correct answer'' (case-sensitive)}\\

\texttt{Examples of correct responses:}\\
\texttt{- Single option: ``2''}\\
\texttt{- Multiple options: ``1,3''}\\
\texttt{- No correct answer: ``No correct answer''}\\

\texttt{Final Reminder:}\\
\texttt{When facts are provided, base your answer primarily on the facts. When facts are empty, you can use basic reasoning about the question and options, but be conservative: only select an option if it clearly and unambiguously satisfies the question's requirements. If there is any significant uncertainty, ambiguity, or if multiple options seem equally valid or if no option clearly fits, return ``No correct answer''. It is better to conservatively return ``No correct answer'' than to incorrectly select an option.}\\

\texttt{Answer:}
\end{tcolorbox}

\begin{tcolorbox}[
    colback=orange!5!white,
    colframe=orange!75!black,
    title=Prompt template for Judging Datasets
]
\scriptsize
\ttfamily
\setlength{\parskip}{0pt}

\texttt{Question: \{conclusion\}}\\
\texttt{Facts: \{facts\}}\\
\texttt{Task: Determine which option(s), if any, are logically supported by the given facts.}\\[1pt]

\texttt{Critical Assumptions (Do NOT violate):}\\
\texttt{1. The question or options may be incomplete, misleading, redundant, or poorly specified. Each question must be evaluated independently.}\\
\texttt{2. There may be zero, one, or multiple correct options. All cases are valid and should be handled appropriately.}\\
\texttt{3. Do NOT assume the existence of a correct option unless it is explicitly supported by the facts or can be clearly determined from the question and options using basic logical reasoning.}\\
\texttt{4. Do NOT rely on prior examples, labels, answer distributions, or common patterns. Each question is independent and should be evaluated on its own merits.}\\
\texttt{5. Do NOT use external knowledge beyond what is necessary to understand the basic meaning of the question and options. Stick to what can be determined from the provided information.}\\
\texttt{6. If facts are provided, they take precedence over any general knowledge. Base your answer primarily on the facts.}\\[1pt]

\texttt{Reasoning Procedure (Follow Step-by-Step):}\\
\texttt{Step 1: Examine the facts (if provided):}\\
\texttt{- If facts are provided, use them as the primary source of information.}\\
\texttt{- Evaluate each option against the facts to determine if it is supported, contradicted, or unsupported.}\\
\texttt{- Supported: facts explicitly state it or logically entail it.}\\
\texttt{- Contradicted: facts explicitly state the opposite or logically entail the opposite.}\\
\texttt{- Unsupported: facts provide no relevant information.}\\

\texttt{Step 2: Handle empty facts (if facts are empty or state ``No additional context provided''):}\\
\texttt{- You may use basic logical reasoning and general knowledge, but be conservative.}\\
\texttt{- Classification (e.g., ``Select the bird''): answer from question+options if clear.}\\
\texttt{- Comparison: analyze given examples and compare with options.}\\
\texttt{- Knowledge-based (e.g., capital of France): if not derivable from options, return ``No correct answer''.}\\
\texttt{- Critical test: if none fits clearly, or all seem equally valid/invalid, return ``No correct answer''.}\\

\texttt{Step 3: Evaluate each option systematically:}\\
\texttt{- For each option, decide: Supported/Entailed, Contradicted, or Unsupported/Ambiguous.}\\
\texttt{- Multiple options may be correct only if logically compatible.}\\

\texttt{Step 4: Make the final decision:}\\
\texttt{- Exactly one clearly supported $\rightarrow$ select it.}\\
\texttt{- Multiple clearly supported and compatible $\rightarrow$ select all.}\\
\texttt{- None clearly supported or significant ambiguity $\rightarrow$ ``No correct answer''.}\\
\texttt{- When in doubt, return ``No correct answer'' rather than guessing.}\\[1pt]

\texttt{Options:}\\
\texttt{\{options\_text\}}\\[1pt]

\texttt{Response Format (Strictly Follow):}\\
\texttt{- Single correct option: output only its number (e.g., ``2'')}\\
\texttt{- Multiple correct options: output all numbers separated by commas, no spaces (e.g., ``1,3'')}\\
\texttt{- No correct options: output exactly ``No correct answer'' (case-sensitive)}\\

\texttt{Examples of correct responses:}\\
\texttt{- Single option: ``2''}\\
\texttt{- Multiple options: ``1,3''}\\
\texttt{- No correct answer: ``No correct answer''}\\[1pt]

\texttt{Final Reminder:}\\
\texttt{When facts are provided, base your answer primarily on the facts. When facts are empty, use only basic conservative reasoning: select an option only if it clearly and unambiguously satisfies the question. If there is any significant uncertainty or ambiguity, return ``No correct answer''.}\\

\texttt{Answer:}
\end{tcolorbox}

\end{document}